\documentclass[10pt,twocolumn,letterpaper]{article}

\usepackage{wacv}
\usepackage{times}
\usepackage{epsfig}
\usepackage{graphicx}
\usepackage{amsmath}
\usepackage{amssymb}

\def\wacvPaperID{422} % Enter the WACV Paper ID here

\wacvfinalcopy % *** Uncomment this line for the final submission

\ifwacvfinal
\def\assignedStartPage{1} % *** Enter the assigned starting page number (instead of 9876)
\fi

\ifwacvfinal
\usepackage[breaklinks=true,bookmarks=false]{hyperref}
\else
\usepackage[pagebackref=true,breaklinks=true,colorlinks,bookmarks=false]{hyperref}
\fi

\usepackage{enumitem}
\usepackage{multirow}
\usepackage{algorithm,algorithmic}
\usepackage[table,x11names]{xcolor}
\usepackage{amsmath,amsfonts,amssymb,amsbsy}

\usepackage[frozencache=true,cachedir=_minted-main]{minted}
\colorlet{shadecolor}{gray!20}
\usepackage[accsupp]{axessibility} % Improves PDF readability for those with disabilities.

\ifwacvfinal
\else
\fi

\begin{document}

%%%%%%%%% TITLE
\title{edge--SR: Super--Resolution For The Masses}

\author{Pablo Navarrete Michelini,\quad Yunhua Lu,\quad Xingqun Jiang\\
BOE Technology Group Co., Ltd.
}

\maketitle
%\thispagestyle{empty}

%%%%%%%%% ABSTRACT
\begin{abstract}
    Classic image scaling (e.g. bicubic) can be seen as one convolutional layer and a single upscaling filter. Its implementation is ubiquitous in all display devices and image processing software. In the last decade deep learning systems have been introduced for the task of image super--resolution (SR), using several convolutional layers and numerous filters. These methods have taken over the benchmarks of image quality for upscaling tasks. Would it be possible to replace classic upscalers with deep learning architectures on edge devices such as display panels, tablets, laptop computers, etc.? On one hand, the current trend in Edge--AI chips shows a promising future in this direction, with rapid development of hardware that can run deep--learning tasks efficiently. On the other hand, in image SR only few architectures have pushed the limit to extreme small sizes that can actually run on edge devices at real--time. We explore possible solutions to this problem with the aim to fill the gap between classic upscalers and small deep learning configurations. As a transition from classic to deep--learning upscaling we propose edge--SR (eSR), a set of one--layer architectures that use interpretable mechanisms to upscale images. Certainly, a one--layer architecture cannot reach the quality of deep learning systems. Nevertheless, we find that for high speed requirements, eSR becomes better at trading--off image quality and runtime performance. Filling the gap between classic and deep--learning architectures for image upscaling is critical for massive adoption of this technology. It is equally important to have an interpretable system that can reveal the inner strategies to solve this problem and guide us to future improvements and better understanding of larger networks.
\end{abstract}

%%%%%%%%% BODY TEXT
\section{Introduction}
\begin{figure}
    \centering
    \begin{minipage}[b]{\linewidth}
        \centerline{\includegraphics[width=\linewidth]{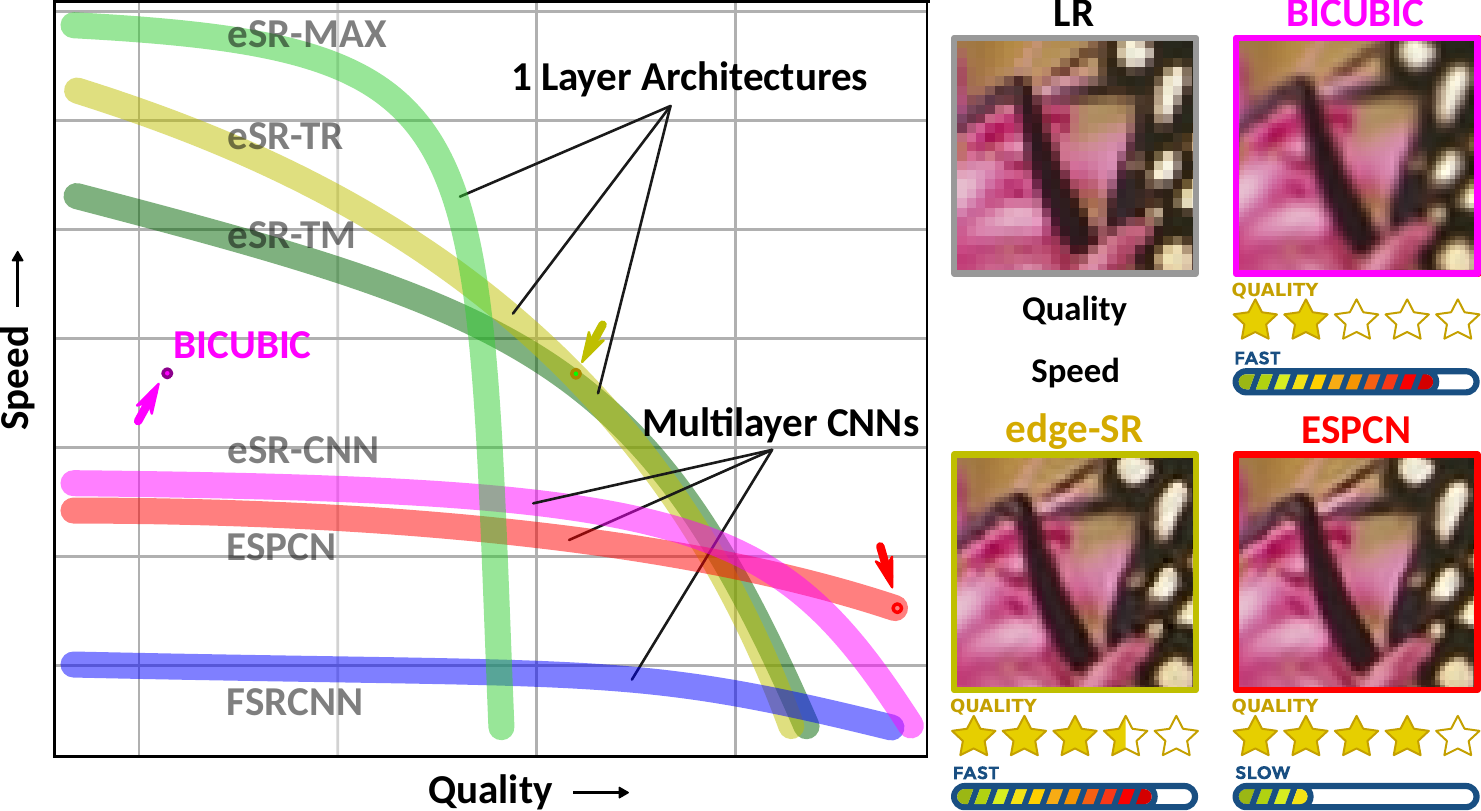}}
    \end{minipage}
\caption{Trade--off pattern observed in our experiments. A standard upscaler (e.g. bicubic) is fast and can be deployed in any display hardware. Multilayer CNNs reach better quality but show sharp loss in quality when their size is reduce to make them faster. We propose edge--SR (eSR) one layer architectures that can adjust the size to reach the best possible quality for a display hardware.}
\label{fig:teaser}
\end{figure}

A market is growing rapidly and steadily to provide so--called Edge--AI chips that will be able to spread the success of deep--learning systems to edge devices \cite{james2021and, li2019artificial, chen2019deep}. This is a massive market that includes phones, tablets and high resolution TV displays, among others. For some applications the success is guaranteed, such as image classification or object detection, where input images are relatively small (e.g. $256\times 256$) and the output data is low dimensional (e.g. labels or bounding boxes). For other applications such as recovering a high--resolution image from a small--resolution image, also known as image super--resolution (SR), the future is less certain since both input and output images can contain a large amount of data. Consider upscaling images from Full--HD to 4K resolution in TV displays for example. The input layer needs to handle $2$ megapixels and the output layer needs to deliver $8$ megapixels at a rate of at least $24$ frames per second. Interestingly, upscaling with small factors (e.g. $2\times$) is both the easiest problem for networks to fix, typically requiring less number of parameters to learn, and at the same time the most difficult solution to deploy. The latter is due to the fact that display devices have a fixed output resolution. For small upscaling factors the input images are still large and demand higher input throughput compared to higher upscaling factors, where input images get smaller and smaller. Small upscaling factors are also of primary concern in applications since they are the most critical technology for transitions between current and new standards (e.g. FHD to 4K, 4K to 8K, etc). Thus, the problem of image SR becomes both more interesting and more challenging given extreme performance constraints.

\textbf{History of SR}. Standard upscaler algorithms, such as linear or bicubic upscalers, apply a low--pass filter on a high resolution image created by inserting zeros between adjacent pixels in the low resolution~\cite{JGProakis_2007a,SMallat_1998a}. Modern tensor processing frameworks (e.g. Pytorch, Tensorflow, etc.) implement this process using a so--called \emph{strided transposed convolutional layer} with a single filter per input channel. More advanced upscalers have followed geometric principles to improve image quality. For example, \emph{edge--directed interpolation} uses adaptive filters to improve edge smoothness \cite{VRAlgazi_1991a,XLi_2001a}, or \emph{bandlet} methods use both adaptive upsampling and filtering \cite{SMallat_2007a}. Later on, machine learning has been able to use examples of pristine high--resolution images to learn a mapping from low--resolution \cite{SCPark_2003a}. The rise of deep--learning and convolutional networks in image classification tasks \cite{YLeCun_2015a} quickly saw a series of important improvements. Many of these improvements followed the progress in network architectures for image classification, as seen for example with CNNs applied in SRCNN~\cite{Dong_2014a}, ResNets~\cite{he2016deep} applied in EDSR~\cite{Lim_2017_CVPR_Workshops}, DenseNets~\cite{huang2017densely} applied in RDN~\cite{zhang2018residual}, attention~\cite{hu2018squeeze-and-excitation} applied in RCAN~\cite{zhang2018rcan}, non--local attention~\cite{wang2018non-local} applied in RNAN~\cite{zhang2019residual}, and swin transformers~\cite{liu2021Swin} applied in SwinIR~\cite{liang2021swinir}.

\textbf{Real--time SR}. The first deep learning system proposed for image SR, namely SRCNN~\cite{Dong_2014a}, used a relatively small number of parameters ($60k$) and became a suitable candidate for edge devices. Soon after, FSRCNN~\cite{Dong_2016a} realized that significant improvements in quality and performance can be achieved by performing computations at low resolution. They proposed a \emph{short} configuration using $4k$ parameters in a sequence of $4$ convolutional layers, plus a final strided transposed convolution to perform upscaling, reaching real--time performance for small resolutions. The next major progress towards real--time applications was made by ESPCN~\cite{shi2016real} that made popular the application of pixel--shuffle layers, multiplexing several network channels to form higher resolution outputs~\cite{PNavarrete_2016a, PNavarrete_2017a}. They proposed a configuration using $20k$ parameters and $3$ convolutional layers with all computations performed at low resolution. Both FSRCNN and ESPCN left a strong mark on future image SR research that very often performs computations at low resolution and use pixel--shuffle layers. Nevertheless, the research clearly shifted to networks of larger sizes that can achieve much better quality. But large networks that contain several million parameters, for example EDSR~\cite{Lim_2017_CVPR_Workshops} (combining ResNets and pixel--shuffle), are currently unable to reach the throughput needed for real--time applications on edge devices. Several so--called lightweight networks have been proposed for middle ground applications~\cite{zhang2020aim,lee2019mobisr,liu2021splitsr,wu2021trilevel,ayazoglu2021extremely,ignatov2021real}. Typical lightweight networks use hundred of thousands parameters and are still beyond the capabilities of real--time applications on edge devices.

\textbf{The Problem}. Despite the promising advances in technology, the challenge of image SR for edge devices remains largely unresolved. One might expect Edge--AI chips to get faster and cheaper but standards also evolve to make problems more difficult (e.g. BT.2020~\cite{MSugawara_2014a}) with more pixels, higher bit depths, higher framerates, etc.. Thus, the success of AI chips to deploy image SR technologies and reach massive markets strongly depends on better algorithm solutions. The major challenge is how to simplify network structures all the way down to reach performance levels comparable to those of classic non--adaptive upscalers. A classic $2\times$ bicubic, doubling the horizontal and vertical resolution, can be implemented using a transposed convolutional layer with a single filter using $121$ parameters. We can think of this as the simplest possible network configuration for image SR. A configuration that is interpretable in the sense that we understand what the interpolation filter values represent. Our main task here is to explore the landscape between classic upscaling on one hand, and small deep--learning systems on the other hand, in order to provide practical solutions for the current state of applications in edge devices.

\textbf{Towards a solution}. Exploring different configurations for existing networks, such as FSRCNN and ESPCN, is a straightforward and necessary task to undertake. But we propose to move a step further, introducing a minimal set of architectures, \textbf{edge--SR (eSR)}, that can perform image SR even with a single convolutional layer. We explore both a straightforward 1--layer Maxout network (eSR-MAX) as well as self--attention strategies (eSR-TM and eSR-TR) that provide a semi--classical interpretation. The latter approaches use a single layer both to detect local patterns (e.g. edges or textures) as well as to generate candidate upscale solutions. Generally speaking, the detection mechanism estimates the probability of the best upscale solution and it is used to compute a weighted average of the candidate output images that gives the final output. We will show how to implement this solution efficiently using standard deep learning modules that can run on AI chips.

\textbf{Contributions}. Our major contributions include:
\begin{itemize}[leftmargin=*]
    \item The \textbf{proposal} of several one--layer architectures that strive for simplicity to fill the gap between classic and deep learning upscalers.
    \item An \textbf{exhaustive search} among $1,185$ network models, including different configurations of eSR, FSRCNN, and ESPCN. Each architecture was trained under identical conditions and tested for speed, power consumption and image quality. The results allows us to visualize the trade--off between image quality and runtime performance that is critical for our purpose. Figure \ref{fig:teaser} shows the general pattern observed in our results. We found that different architectures show very different balance in the trade--off between speed and image quality. Multi--layer networks (deep learning) show a strong advantage at low speed and high quality, and our proposed one--layer solutions show a clear advantage at high speed requirements.
    \item The \textbf{interpretation and analysis} of strategies learned by self--attention in one--layer architectures. We provide a novel interpretation of the self--attention mechanism based on the simple principles of template matching and classic upscaling. Here, training results indicate that one--layer networks do not use smooth upscaling kernels and rely mostly on independent sub--pixel solutions.
\end{itemize}

These results may bring about the following \textbf{future impact}: 1) the possibility of image SR systems that can be massively deployed on edge devices, 2) a better understanding of the internal learning mechanisms of small network architectures, and 3) a better appreciation of the trade--off between image quality and runtime performance for future applications and research.

\begin{figure}
  \centering
  \centerline{\includegraphics[width=.8\linewidth]{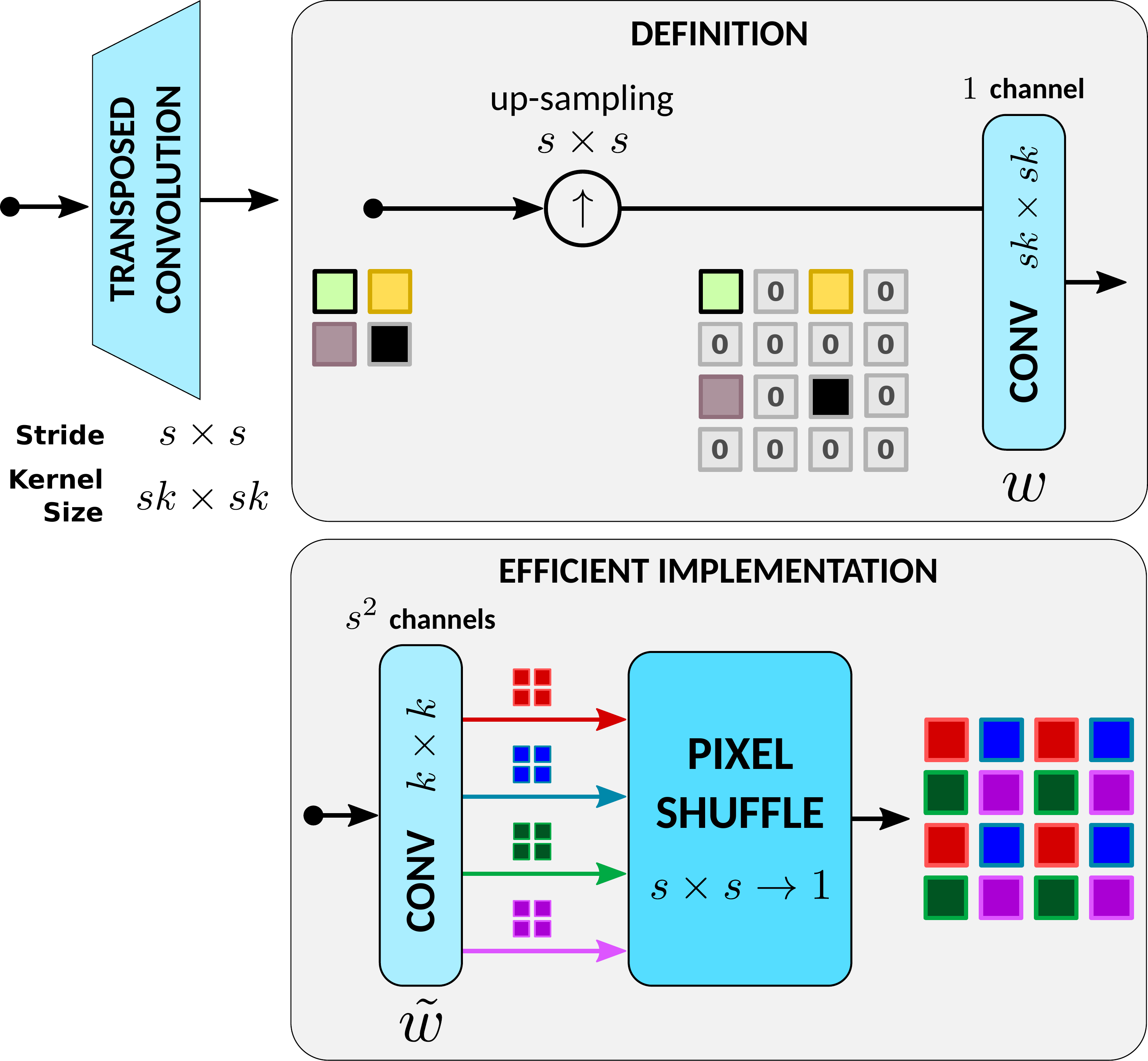}}
  \caption{Classic $s\times s$ image upscaling is performed by a transposed convolutional layer. An efficient implementation splits the filter into $s^2$ smaller filters that work at LR. The final output is obtained by multiplexing the $s^2$ channels using a pixel--shuffle layer.}
  \label{fig:classic}
\end{figure}

\section{Super--Resolution for Edge Devices}

\textbf{Classical.}
Image \emph{upscaling} and \emph{downscaling} refer to the conversion of low resolution (LR) images to high resolution (HR) and vice versa. These two processes are closely related. The simplest way to downscale an image from HR to LR is known as \emph{pooling} or \emph{downsample}. The process of downsample uniformly drops pixels in both horizontal and vertical directions. The problem with such downscalers is that groups of high and low frequency components of the HR image can end up in the same low frequency component at LR, leading to well known \emph{aliasing} artifacts~\cite{JGProakis_2007a,SMallat_1998a}. To avoid this problem a classic \emph{linear downscaler} first removes high frequencies using an \emph{anti--aliasing}  low--pass filter and then downsamples the image. This process is implemented in tensor processing frameworks with \emph{strided convolutional layers} where the kernel or weight parameters correspond to the low--pass filter coefficients. The process of classic \emph{linear upscaling} corresponds to the transposed of the downscaling linear transformation and it is illustrated in Figure \ref{fig:classic}. The transposition reverts the `filter--then--downsampling' operation into an `upsampling--then--filter' operation where the \emph{upsampling} increases the resolution of an image by inserting zeros between LR pixels. The upsampling introduces high frequencies that are removed by a so--called \emph{interpolation} filter with coefficients $w$. The interpolation filter is the transposed of the anti--aliasing filter, typically identical because most upscalers are symmetric. Tensor processing frameworks implement this process using \emph{strided transposed convolutional layers}.

The upscaling definition in Figure \ref{fig:classic} is clearly inefficient as the upsampling introduces many zeros that will waste resources when multiplied by filter coefficients. A very well know optimization, widely used in practical implementations of classic upscalers is to split or demultiplex the interpolation filter from size $sk\times sk$ in Figure \ref{fig:classic} to $s^2$ so--called \emph{efficient} filters of size $k\times k$ working at LR~\cite{JGProakis_2007a,SMallat_1998a}. The outputs of the $s^2$ filters are then multiplexed by a pixel--shuffle operation to obtain the upscaled image, as illustrated in Figure \ref{fig:classic}. Let $\tilde{w}_i\in\mathbb{R}^{k\times k}$, with $i=1,\ldots,s^2$, be the coefficients of the efficient filters. The interpolation filter can then be recovered  by multiplexing the efficient coefficients back to their original place. This is,
\begin{equation}
  w = \text{Pixel--Shuffle}_{s\times s}(\tilde{w}_i, i=1\ldots,s^2) \;. \label{eq:shuffle_filters}
\end{equation}

In our experiments we will compare different architectures including a bicubic upscaler. In order to remove implementation advantages we implemented the upscaler using the efficient implementation in Figure \ref{fig:classic}. We used standard bicubic interpolation filter coefficients and verified that we obtain the same outputs as other software implementations up to floating point precision.

\begin{figure}
  \centering
  \begin{minipage}[b]{.8\linewidth}
    \includegraphics[width=\linewidth]{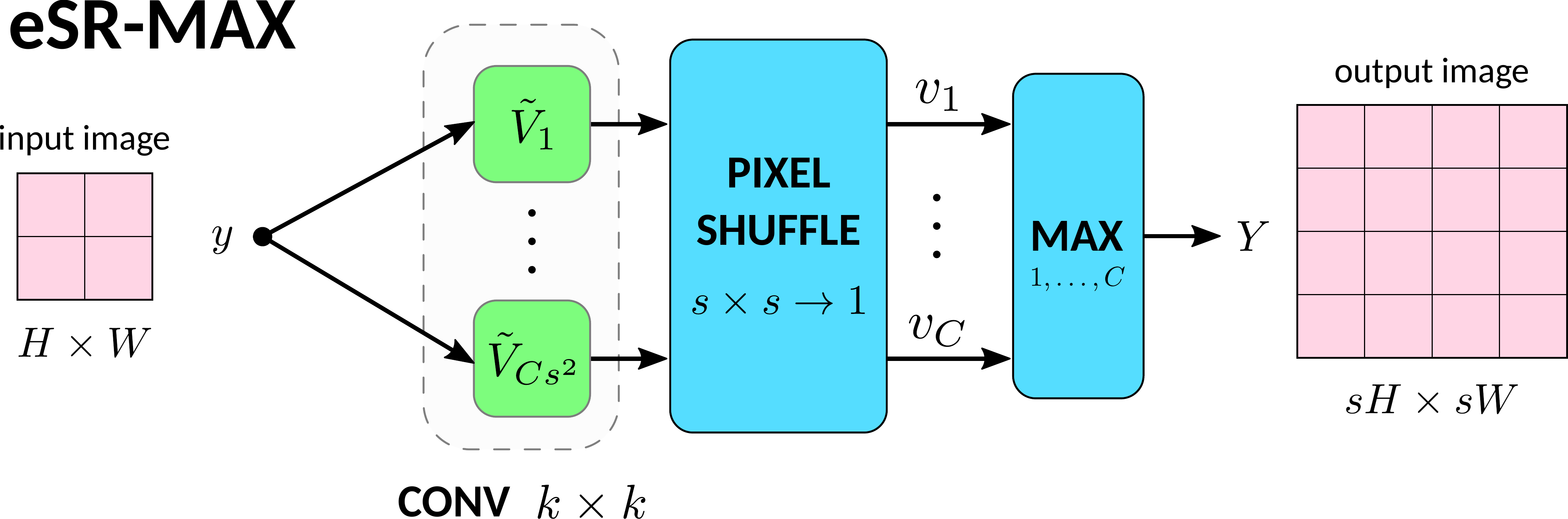}
  \end{minipage}
  \caption{Diagram of edge--SR Maximum. Uses one convolutional layer followed by a pixel--shuffle multiplexer and a non--linear module that chooses the maximum pixel value among all filters.}
  \label{fig:esr-max}
\end{figure}

\begin{figure*}
  \begin{minipage}[b]{1.\linewidth}
    \centering
    \centerline{\includegraphics[width=\linewidth]{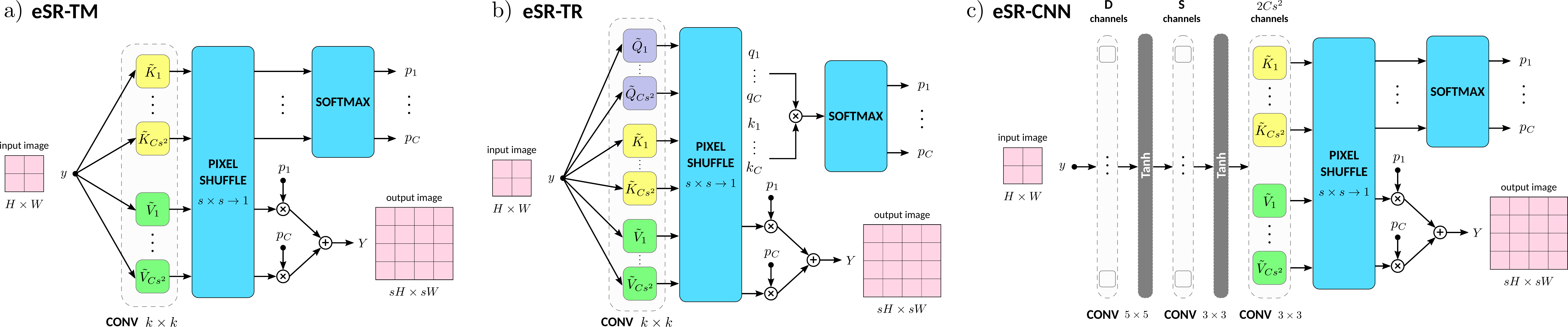}}
  \end{minipage}
  \caption{Diagrams of edge--SR architectures using self--attention: a) edge--SR Template Matching (esR--TM) runs both template matching ($K$ filters) and upscaling modules ($V$ filters) using a single convolutional layer (see Figures \ref{fig:schematic} and \ref{fig:esr_interp} for details on this interpretation), b) edge--SR TRansformer (eSR--TR) uses two sets of \emph{query} ($Q$) and \emph{key} ($K$) filters to estimate the best upscaler model, and c) edge--SR CNN (eSR--CNN) starts with a multilayer network following ESPCN and ends with a eSR--TM module.}
  \label{fig:esr}
\end{figure*}

\begin{algorithm*}[t]
    \centering
    \begin{tabular}{ll}
        \textbf{\emph{eSR--MAX}}$(y,C,k,s)$: & \textbf{\emph{eSR--TM}}$(y,C,k,s)$: \\

        \resizebox{.455\textwidth}{!}{
            \begin{minipage}{.5\textwidth}
                \vspace{-0.1in}
                \begin{algorithmic}[1]
                    \REQUIRE Integer $C> 1, k>1, s>1$.

                    \STATE $Y = \text{Max}_{1\rightarrow C}\;\text{Pixel--Shuffle}_{s\times s}\big( \text{Conv}^{k\times k}(y)\big)$
                \end{algorithmic}
        \end{minipage}
        }

        &

        \resizebox{.455\textwidth}{!}{
            \begin{minipage}{.5\textwidth}
                \begin{algorithmic}[1]
                    \REQUIRE Integer $C> 1, k>1, s>1$.

                    \STATE $f = \text{Pixel--Shuffle}_{s\times s}\big( \text{Conv}^{k\times k}(y) \big)$
                    \STATE $Y = \sum_{1\rightarrow C}\big(f_{C+1 \;\rightarrow\; 2\cdot C} \otimes \text{SoftMax}(f_{1 \;\rightarrow\; C})\big)$
                \end{algorithmic}
        \end{minipage}
        }

        \\ \\
        \textbf{\emph{eSR--TR}}$(y,C,k,s)$: & \textbf{\emph{eSR--CNN}}$(y,C,D,S,s)$: \\

        \resizebox{.455\textwidth}{!}{
            \begin{minipage}{0.5\textwidth}
                \begin{algorithmic}[1]
                    \REQUIRE Integer $C> 1, k>1, s>1$.

                    \STATE $f = \text{Pixel--Shuffle}_{s\times s}\big( \text{Conv}^{k\times k}(y) \big)$
                    \STATE $p = \text{SoftMax}(f_{1 \;\rightarrow\; C} \otimes f_{C+1 \;\rightarrow\; 2\cdot C})$
                    \STATE $Y = \sum_{1\rightarrow C}\big(f_{2\cdot C+1 \;\rightarrow\; 3\cdot C} \otimes p \big)$
                \end{algorithmic}
            \end{minipage}
        }

        &

        \resizebox{.455\textwidth}{!}{
            \begin{minipage}{0.5\textwidth}
                \begin{algorithmic}[1]
                    \REQUIRE Integer $C> 1, D>1, S>1, s>1$.

                    \STATE $f = \text{Pixel--Shuffle}_{s\times s}\;\circ\;\text{Conv}^{5\times 5}\;\circ\;\text{Tanh}\;\circ\;\text{Conv}^{3\times 3}\;\circ\;\text{Tanh}\;\circ\;\text{Conv}^{3\times 3}(y)$
                    \STATE $Y = \sum_{1\rightarrow C}\big(f_{C+1 \;\rightarrow\; 2\cdot C} \otimes \text{SoftMax}(f_{1 \;\rightarrow\; C})\big)$
                \end{algorithmic}
            \end{minipage}
        }
    \end{tabular}
    \caption{edge Super--Resolution (eSR). \hspace{.1in} \textbf{Input :} $y$ (1--channel). \hspace{.1in} \textbf{Output :} $Y$ (1--channel).}
    \label{alg:esr}
\end{algorithm*}

\textbf{Maxout.}
Our first proposal is edge--SR Maximum (eSR--MAX). This is an attempt to obtain the fastest solution from a single convolutional layer that outputs several upscaled candidates. A quick decision is made by choosing the maximum value across all channels as shown in Figure \ref{fig:esr-max}. This corresponds to a particular case of a Maxout network \cite{goodfellow2013maxout}.

\begin{figure}
  \begin{minipage}[b]{1.0\linewidth}
    \centering
    \centerline{\includegraphics[width=\linewidth]{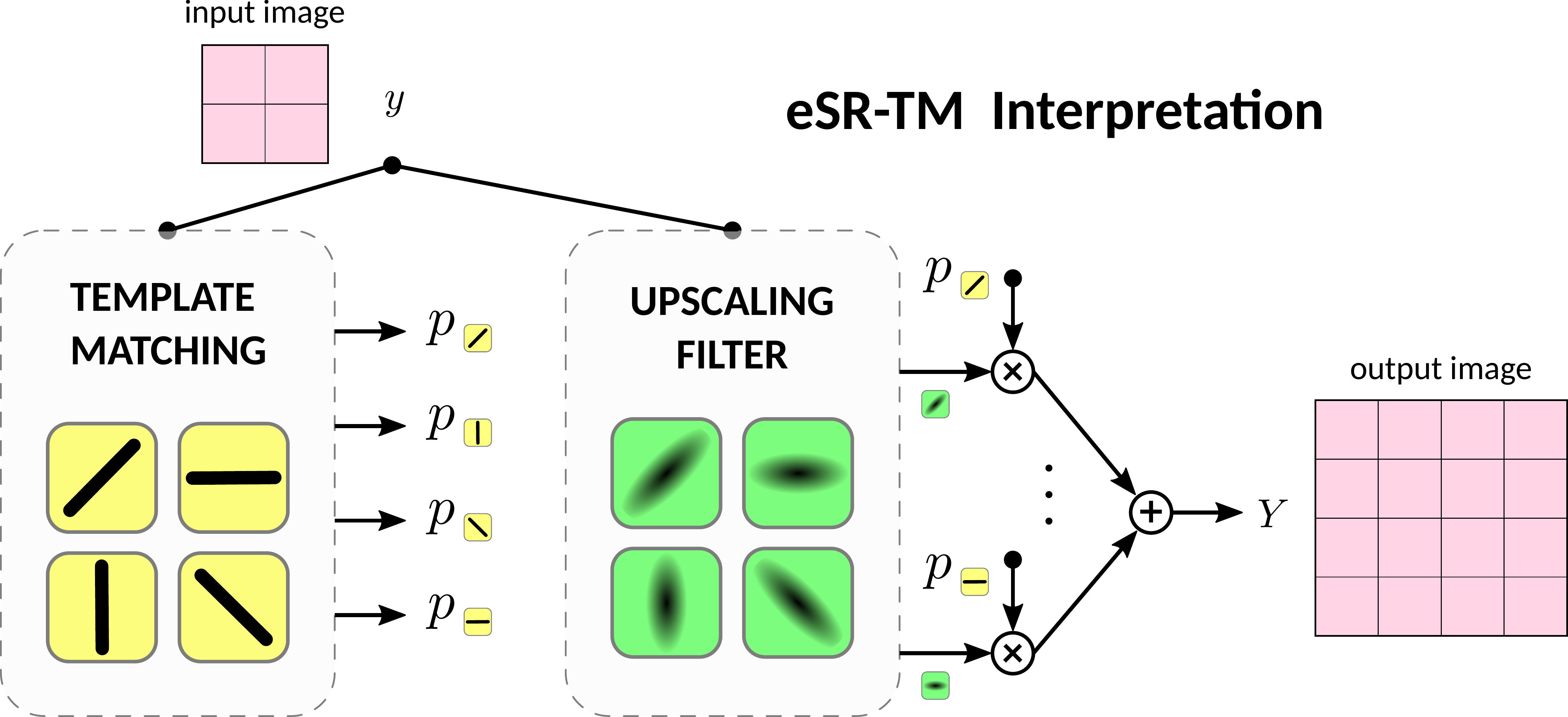}}
  \end{minipage}
\caption{Interpretation of edge--SR Template Matching. A template matching module computes the probability of finding one of the template image features learned from examples. The input image is upscaled using a set of different upscalers also learned by examples. The output is the expected value computed by the weighted average of upscale images and template probabilities.}
\label{fig:schematic}
\end{figure}

\textbf{Self--Attention.}
Our second proposal is edge--SR Template Matching (eSR--TM) that follows a semi--classical strategy. The basic idea is explained in Figure \ref{fig:schematic}. First, a template matching module detects patterns (e.g. edge directions) and gives us the probability for each pattern. This is achieved by: first, use matching filter coefficients that resemble the pattern, and second, normalize pixel values across channels to represent the probability of each template. A set of upscale images are computed at the same time for each one of the patterns. Since both the matching and the upscaling filters follow the same patterns, \emph{we expect the filter coefficients to look similar} as displayed in Figure \ref{fig:schematic} for the case of edge patters. Thus, we can verify if an eSR--TM configuration learned to perform template matching by checking the correlations between filter coefficients. The optimal prediction for the output image is the expected value over all templates. Thus, the probabilities are used to compute the expected value by weighing the solution of different upscalers that when combined give the final output.

Figure \ref{fig:esr}.a shows the diagram of the efficient implementation of this idea using $C\in\mathbb{N}^+$ templates. In this efficient implementation of a transposed convolution the $C$ matching filters $K$ split into $Cs^2$ efficient filters $\tilde{K}$, before multiplexing with pixel--shuffle. We can always get the interpolation filters $K$ from $\tilde{K}$ using equation \eqref{eq:shuffle_filters}. The outputs of the filters are then normalized among all the channels using a softmax module. This gives us the pixel--wise probabilities:
\begin{equation}
p_i = e^{K_i \circledast (y\;\uparrow\; s)}/\sum_{j=1}^{C} e^{K_j \circledast (y\;\uparrow\; s)} \;,
\end{equation}
where $i=1,\ldots,C$, $\circledast$ is the convolution operator, $\uparrow$ refers to the upsampling operation defined in Figure \ref{fig:classic}. The same convolutional layer in Figure \ref{fig:esr}.a runs $Cs^2$ efficient filters $\tilde{V}$ to get $C$ high resolution candidates after pixel--suffle. The final luminance HR output image $Y$ is given by:
\begin{equation}
  Y=\mathbb{E}\left[V_i \circledast (y\uparrow s)\right] = \sum_{i=1}^C p_i \otimes (V_i \circledast (y\uparrow s)) \;,
\end{equation}
where $\otimes$ represent a Hadamard (or pixel--wise) product.

The eSR--TM system is essentially a self--attention module, except for the pixel--shuffle layer and the sum over all channels in the last stage. These two differences are significant since: first, they embed the upscaling process within the attention module, and second, they make explicit use of probabilities to compute an expected value thus providing a clear interpretation of this module.

Our third proposal is edge--SR TRansformer (eSR--TR) that uses the popular \emph{transformer} self--attention module from \cite{vaswani2017attention}. Figure \ref{fig:esr}.b shows the efficient implementation of this system. Here, the matching filters from eSR--TM are replaced by two sets of \emph{query} ($Q$) and \emph{key} ($K$) filters to estimate the probabilities. This changes the template matching interpretation of eSR-TM, using a rank--1 quadratic form with $Q$ and $K$ filters instead of a single template matching filter. The purpose of this arquitecture is to test any advantage that this change could bring given the increasing popularity and success of this module in recent research.

The code for all eSR systems is given in Algorithm \ref{alg:esr}.

\begin{figure}
  \begin{minipage}[b]{1.0\linewidth}
    \centering
    \centerline{\includegraphics[width=\linewidth]{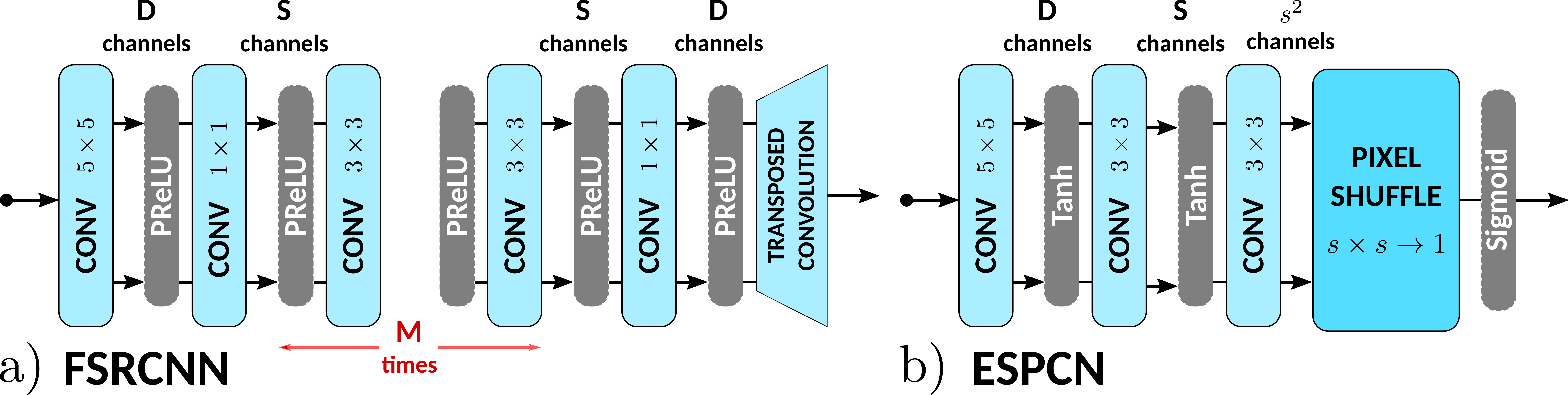}}
  \end{minipage}
  \caption{Deep--learning architectures selected for experiments: a) FSRCNN from \cite{Dong_2016a}, and b) ESPCN from \cite{shi2016real}.}
  \label{fig:deeplearning}
\end{figure}

\textbf{Deep--Learning.}
We consider FSRCNN~\cite{Dong_2016a} and ESPCN~\cite{shi2016real} as candidate deep learning architectures for image SR on edge devices. Figure \ref{fig:deeplearning} shows the detail structure of FSRCNN and ESPCN network architectures. In comparison, FSRCNN uses more layers (at least $5$) and smaller number of channels per layer than ESPCN. Another difference is the upscaling strategy, with FSRCNN using a strided transposed convolution and ESPCN using pixel--shuffle. According to classic interpolation theory these two approaches are equivalent as shown in Figure \ref{fig:classic} (see also \cite{JGProakis_2007a,SMallat_1998a}), but implementations can be different. Tensor processing frameworks typically implement transposed convolution using the gradient of a convolutional layer\cite{parr2018matrix}, based on the vector calculus property for gradients of linear transformations: $\nabla_x(Ax+b) y = A^T y$. This very different approach might lead to differences in performance.

Finally, we also propose the edge--SR CNN (eSR--CNN) architecture in Figure \ref{fig:esr}.c and Algorithm \ref{alg:esr}. This is simply an extension of the single convolutional layer in eSR--TM into a multi--layer structure identical to ESPCN. Here, the purpose is to test if ESPCN, that achieves better results compared to FSRCNN in our tests, can be improved by using a self--attention module to upscale.

\section{Experiments}

\textbf{Models.} Candidate models for test evaluations include: bicubic, FSRCNN, ESPCN and eSR. From these, the bicubic classic upscaler is the only one without hyper--parameters and fixed configuration that do not require training. For other architectures we need to train a model for each set of hyper--parameters. Table \ref{tab:parameters} shows the list of hyper--parameters chosen for our experiments. These include default settings of FSRCNN and ESPCN as well as configurations with very small number of parameters. Our model pool includes a total of $1,185$ models to evaluate.

\begin{table}
    \caption{Set of hyper--parameters used to create a pool of $1,185$ models that were trained and tested in our experiments.}
    \label{tab:parameters}
    \medskip
    \centering
    \resizebox{1.\linewidth}{!}{
    \begin{tabular}{lrl}
        \hline
        Bicubic & \textbf{Total :} & \textbf{1} model per scale factor. \\
        \hline
        \multirow{4}{*}{eSR}
        & $C$ :  &  1, 2, 3, 4, 5, 6, 7, 8, 9, 10, 11, 12, 13, 14, 15, 16. \\
        & $k$ :  &  3, 5, 7. \\
        & Type :  &  Maximum (MAX), Template Matching (TM), Transformer (TR). \\
        & \textbf{Total :} & \textbf{144} models per scale factor. \\
        \hline
        \multirow{4}{*}{eSR-CNN}
        & $C$ :  &  2, 4, 6, 8. \\
        & $D$ :  &  1, 3, 5, 7, 9. \\
        & $S$ :  &  3, 6, 9, 12, 15. \\
        & \textbf{Total :} & \textbf{100} models per scale factor. \\
        \hline
        \multirow{4}{*}{FSRCNN}
        & $D$ :  &  6, 19, 32, 44, 56. \\
        & $S$ :  &  1, 3, 6, 9, 12. \\
        & $M$ :  &  1, 4 \\
        & \textbf{Total :} & \textbf{50} models per scale factor. \\
        \hline
        \multirow{3}{*}{ESPCN}
        & $D$ :  &  0, 4, 6, 10, 12, 16, 28, 40, 52, 64. \\
        & $S$ :  &  3, 6, 9, 12, 15, 18, 21, 24, 27, 32. \\
        & \textbf{Total :} & \textbf{100} models per scale factor. \\ \hline
        \rowcolor{lightgray} \multicolumn{2}{r}{Factors :} & $2\times$, $3\times$, $4\times$. \\
        \rowcolor{lightgray} \multicolumn{2}{r}{\textbf{Total :}} & \textbf{1,185} models. \\ \hline
    \end{tabular}
    }
\end{table}

\begin{figure*}
  \centering
  \centerline{\includegraphics[width=\linewidth]{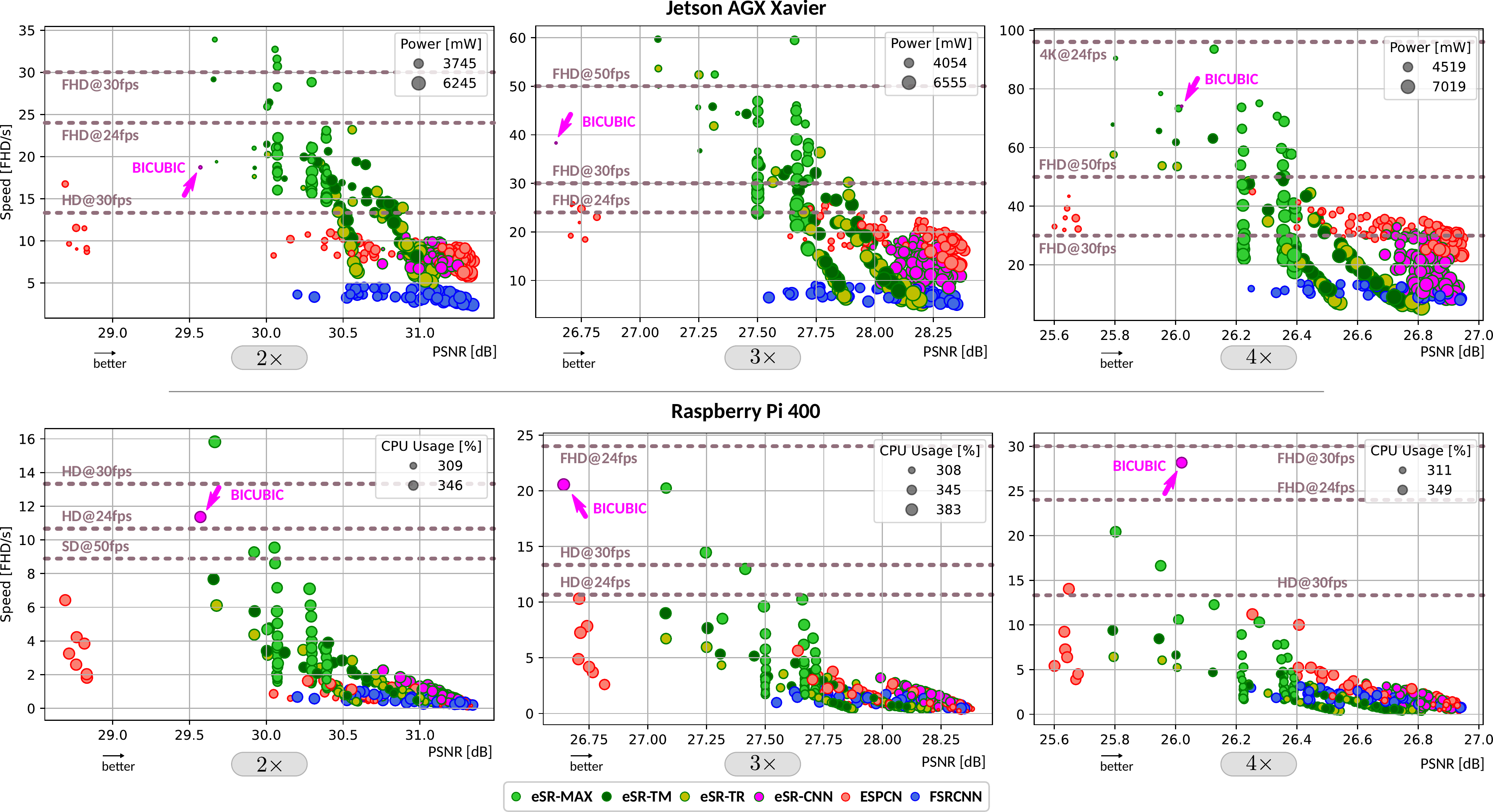}}
  \caption{Scatter plot to compare speed, in number of Full--HD pixels per second, with respect to quality, measured as PSNR for the BSDS--100 dataset. A total of $1,185$ models were identically trained considering different upscaling factors ($2\times$, $3\times$ and $4\times$) and architectures (eSR, ESPCN and FSRCNN). We run all models on edge devices: Jetson AGX Xavier (GPU with 16--bit floating point precision) and Raspberry Pi 400 (CPU with 32--bit floating point precision). Magnified plots with model annotations are provided in Appendix \ref{sec:analysis}.}
\label{fig:psnr_vs_speed}
\end{figure*}

\textbf{Training.} We need to train a total of $1,185$ models that include different scaling factors, network architectures and model hyper--parameters. We trained all these models independently using an identical procedure. We used the General--100 dataset~\cite{Dong_2016a} combined with 91--image dataset~\cite{yang2010image} to extract training patches. For each image in the dataset we randomly cut a HR patch of size $78\times 78$ for $2\times$ and $3\times$ upscaling factors, and $76\times 76$ for $4\times$ factor. The images were converted to grayscale using BT.609 color matrix and downscaled using a standard Bicubic algorithm. We used minibatch size $16$ and trained each model for $25,000$ epochs using a standard mean--square--error (MSE) loss. We started with a learning rate of $10^{-3}$ and reduce it to half once every $3,000$ epochs. We used Adam optimizer~\cite{kingma2014adam} with $\beta_1=0.9$, $\beta_2=0.999$ and $\epsilon=10^{-8}$. We used seven Tesla M40 GPUs for training with the whole process completed in about two months.

\textbf{Measurements.} To test our final models we considered two inference devices: 1) Nvidia Jetson AGX Xavier, an embedded system--on--module (SoM) from the Nvidia AGX Systems family, including an integrated Volta GPU with tensor cores, and 2) a Raspberry Pi 400, an embedded device featuring a quad--core 1.8GHz, 64--bit ARM Cortex CPU processor. The power consumption of the Jetson AGX is set to a 30 Watt profile, while the Raspberry Pi 400 nominal consumption is 15 Watt.

We run each model to output a set of $14$ Full--HD images, downscaling appropriately from randomly selected images of the DIV2K dataset~\cite{Agustsson_2017_CVPR_Workshops}. We use 16--bit floating point precision during inference. For each image we run the model $10$ times to avoid warm--up effects, measuring the minimum CPU and GPU processing time from profiler's data. We computed the speed of a model using the total number of pixels processed (considering only one run per image) divided by the processing time (using the minimum time over each one of the $10$ runs). To make the measurement of speed easier to read we use units of \texttt{[FHD/s]}, this is, number of Full--HD pixels ($1920\times 1080$) per second.

Image quality was measured separately using the standard datasets: Set--5, Set--14\cite{zeyde2010single}, BSDS--100\cite{martin2001database}, Urban--100\cite{huang2015single} and Manga--109\cite{matsui2017sketch-based}. We also measured maximum power consumption for the Jetson AGX and CPU usage for the Raspberry Pi that does not include power sensors.

\textbf{Results.}
Figure \ref{fig:psnr_vs_speed} shows scatter plots to compare speed with respect to image quality, measured as PSNR for the BSDS--100 dataset. Results for other datasets, metrics (SSIM) and devices (GTX 1080 Max--Q) are shown in Appendix \ref{sec:analysis} with similar conclusions. The size of the circles are proportional to the power consumption and CPU usage for the AGX and Raspberry Pi devices, respectively. Finally, Table \ref{tab:imageq} shows detailed results per dataset for a subset of the models selected according to different criteria.

\begin{table*}[ht]
  \caption{Image quality and performance metrics for selected methods among all $1,185$ models trained in our experiments. Values of speed, measured in number of Full--HD pixels per second, and power, in units of Milliwatts, are specific of a Jetson AGX Xavier GPU. Methods are selected based on best speed, PSNR in BSDS--100 dataset, and default configurations. Best results are shown in bold (ignoring bicubic).}
  \label{tab:imageq}
  \centering\medskip
\resizebox{\linewidth}{!}{
  \begin{tabular}{lcclrrcccccccccc}\hline
    \multirow{2}{*}{Algorithm} & \multirow{2}{*}{$s$} & \multirow{2}{*}{Selection} & \multirow{2}{*}{Configuration} & Speed & Power & \multicolumn{2}{c}{Set5} & \multicolumn{2}{c}{Set14} & \multicolumn{2}{c}{BSDS100} & \multicolumn{2}{c}{Urban100} & \multicolumn{2}{c}{Manga109} \\
                             & & & & $\text{[FHD/s]}$ & $\text{[mWatts]}$ & PSNR & SSIM & PSNR & SSIM & PSNR & SSIM & PSNR & SSIM & PSNR & SSIM \\ \hline
    Bicubic & 2 & -- & -- & 19 & 1550 & 33.73 & 0.928 & 30.29 & 0.869 & 29.57 & 0.842 & 26.89 & 0.841 & 30.85 & 0.934 \\
    eSR & 2 & PSNR & CNN: $C=6$, $D=3$, $S=15$ & 8 & 3868 & 36.58 & 0.953 & 32.38 & 0.905 & 31.25 & 0.885 & 29.26 & 0.891 & 35.33 & 0.965 \\
    eSR & 2 & speed & MAX: $k=3$, $C=1$ & \textbf{34} & \textbf{1859} & 33.15 & 0.928 & 30.16 & 0.882 & 29.66 & 0.862 & 26.94 & 0.857 & 30.46 & 0.937 \\
    ESPCN & 2 & default~\cite{shi2016real} & $D=64$, $S=32$ & 6 & 6800 & 36.64 & 0.953 & 32.46 & \textbf{0.907} & 31.32 & \textbf{0.887} & 29.37 & 0.893 & 35.76 & \textbf{0.967} \\
    ESPCN & 2 & PSNR & $D=22$, $S=32$ & 8 & 4945 & 36.70 & 0.953 & \textbf{32.47} & \textbf{0.907} & \textbf{31.35} & \textbf{0.887} & \textbf{29.44} & 0.894 & 35.79 & \textbf{0.967} \\
    ESPCN & 2 & speed & $D=0$, $S=3$ & 17 & 2324 & 29.76 & 0.919 & 28.96 & 0.881 & 28.69 & 0.865 & 26.38 & 0.853 & 27.66 & 0.938 \\
    FSRCNN & 2 & default~\cite{Dong_2016a} & $D=32$, $S=6$, $M=1$ & 4 & 4793 & 36.29 & 0.951 & 32.20 & 0.904 & 31.10 & 0.884 & 28.91 & 0.886 & 35.03 & 0.963 \\
    FSRCNN & 2 & PSNR & $D=56$, $S=12$, $M=4$ & 2 & 5566 & \textbf{36.74} & \textbf{0.954} & 32.45 & \textbf{0.907} & 31.34 & \textbf{0.887} & 29.42 & \textbf{0.895} & \textbf{35.87} & \textbf{0.967} \\
    FSRCNN & 2 & speed & $D=6$, $S=3$, $M=1$ & 5 & 3560 & 35.36 & 0.943 & 31.52 & 0.898 & 30.64 & 0.878 & 28.01 & 0.870 & 33.13 & 0.951 \\
    \noalign{\smallskip}\hline\noalign{\smallskip}
    Bicubic & 3 & -- & -- & 38 & 1705 & 29.24 & 0.849 & 26.73 & 0.757 & 26.64 & 0.720 & 23.84 & 0.717 & 25.87 & 0.838 \\
    eSR & 3 & PSNR & CNN: $C=8$, $D=3$, $S=15$ & 11 & 5873 & 32.75 & \textbf{0.906} & 29.27 & 0.820 & 28.36 & 0.782 & 26.14 & 0.796 & 30.16 & 0.907 \\
    eSR & 3 & speed & TM: $k=3$, $C=1$ & \textbf{60} & \textbf{2632} & 29.77 & 0.853 & 27.31 & 0.780 & 27.08 & 0.748 & 24.31 & 0.742 & 26.56 & 0.851 \\
    ESPCN & 3 & default~\cite{shi2016real} & $D=64$, $S=32$ & 13 & 6027 & 32.73 & 0.905 & 29.26 & \textbf{0.821} & 28.36 & 0.783 & 26.12 & 0.795 & 30.36 & 0.908 \\
    ESPCN & 3 & PSNR & $D=22$, $S=32$ & 16 & 4945 & \textbf{32.77} & \textbf{0.906} & \textbf{29.30} & \textbf{0.821} & \textbf{28.38} & \textbf{0.784} & \textbf{26.15} & \textbf{0.797} & \textbf{30.37} & \textbf{0.909} \\
    ESPCN & 3 & speed & $D=0$, $S=21$ & 26 & 4176 & 31.30 & 0.889 & 28.51 & 0.808 & 27.82 & 0.772 & 25.41 & 0.774 & 28.29 & 0.887 \\
    FSRCNN & 3 & default~\cite{Dong_2016a} & $D=32$, $S=6$, $M=1$ & 8 & 4640 & 32.43 & 0.901 & 29.07 & 0.816 & 28.22 & 0.780 & 25.82 & 0.787 & 29.61 & 0.899 \\
    FSRCNN & 3 & PSNR & $D=56$, $S=12$, $M=4$ & 5 & 5566 & 32.74 & \textbf{0.906} & 29.25 & 0.820 & 28.35 & \textbf{0.784} & 26.10 & \textbf{0.797} & 30.13 & 0.907 \\
    FSRCNN & 3 & speed & $D=6$, $S=1$, $M=1$ & 9 & 3560 & 31.30 & 0.879 & 28.31 & 0.803 & 27.75 & 0.768 & 25.06 & 0.761 & 27.98 & 0.870 \\
    \noalign{\smallskip}\hline\noalign{\smallskip}
    Bicubic & 4 & -- & -- & 74 & 2170 & 28.60 & 0.808 & 26.09 & 0.705 & 26.02 & 0.672 & 23.17 & 0.660 & 24.96 & 0.787 \\
    eSR & 4 & PSNR & CNN: $C=8$, $D=9$, $S=6$ & 13 & 7100 & \textbf{30.62} & 0.860 & 27.48 & 0.751 & 26.93 & 0.714 & 24.42 & 0.718 & 27.27 & 0.845 \\
    eSR & 4 & speed & MAX: $k=3$, $C=2$ & \textbf{94} & 3867 & 28.64 & 0.806 & 26.12 & 0.712 & 26.13 & 0.684 & 23.28 & 0.668 & 25.08 & 0.789 \\
    ESPCN & 4 & default~\cite{shi2016real} & $D=64$, $S=32$ & 23 & 6952 & 30.57 & 0.858 & 27.50 & 0.752 & 26.92 & 0.715 & 24.42 & 0.718 & 27.44 & 0.848 \\
    ESPCN & 4 & PSNR & $D=16$, $S=32$ & 29 & 4640 & 30.59 & 0.859 & \textbf{27.53} & \textbf{0.753} & \textbf{26.95} & 0.715 & 24.43 & 0.719 & \textbf{27.46} & \textbf{0.849} \\
    ESPCN & 4 & speed & $D=1$, $S=3$ & 45 & \textbf{3096} & 28.93 & 0.820 & 26.49 & 0.725 & 26.25 & 0.694 & 23.56 & 0.680 & 25.49 & 0.804 \\
    FSRCNN & 4 & default~\cite{Dong_2016a} & $D=32$, $S=6$, $M=1$ & 12 & 4795 & 30.16 & 0.845 & 27.19 & 0.742 & 26.74 & 0.707 & 24.09 & 0.702 & 26.63 & 0.826 \\
    FSRCNN & 4 & PSNR & $D=44$, $S=12$, $M=4$ & 9 & 5257 & 30.61 & \textbf{0.861} & 27.52 & \textbf{0.753} & 26.94 & \textbf{0.716} & \textbf{24.44} & \textbf{0.721} & 27.40 & \textbf{0.849} \\
    FSRCNN & 4 & speed & $D=6$, $S=1$, $M=1$ & 14 & 3715 & 29.31 & 0.823 & 26.62 & 0.730 & 26.41 & 0.699 & 23.62 & 0.683 & 25.72 & 0.802 \\
    \noalign{\smallskip}\hline\noalign{\smallskip}
  \end{tabular}
}
\end{table*}

\section{Analysis}

\textbf{Trade-off.} The results displayed in Figure \ref{fig:psnr_vs_speed} allow us to fully appreciate the trade--off between image quality and runtime performance. The bicubic upscaler sets the target as we know that it can be massively deployed in display devices at large scale. Between the bicubic upscaler and deep--learning configurations using FSRCNN, ESPCN or eSR--CNN we observe a large empty region. Our proposed edge--SR (eSR) architectures succeeds to fill this gap in edge GPU devices (AGX and also GTX 1080 MaxQ available in Appendix \ref{sec:analysis}) and improve bicubic upscaler both in speed and image quality. In the Raspberry Pi CPU device edge--SR partially succeeds to fill this gap for $2\times$ and $3\times$ upscaling factor and fails at $4\times$ factor where bicubic reaches a better performance. The best results of edge--SR is observed for $2\times$ upscaling factor. The distribution of scatter points in Figure \ref{fig:psnr_vs_speed} for $2\times$ upscaling shows that deep--learning methods are better at image quality, with ESPCN achieving the best speed in the high quality range. Interestingly, eSR--CNN does not improve ESPCN at high quality and barely improves it at high speed, despite using the same multi--layer configuration. eSR--MAX shows the best performance at high speeds but it is unable to make significant improvements in image quality. eSR--TM and eSR--TR show the best performance at intermediate speed and image quality. They perform very similar with a slight but not conclusive advantage of eSR--TR on GPU devices. FSRCNN shows the worst performance at $2\times$ factor with no improvements in speed as image quality decreases. One possible reason for this result is that the higher network depth of FSRCNN might become a disadvantage at $2\times$ where large receptive fields are unnecessary.

\begin{figure}
    \centering
    \includegraphics[width=\linewidth]{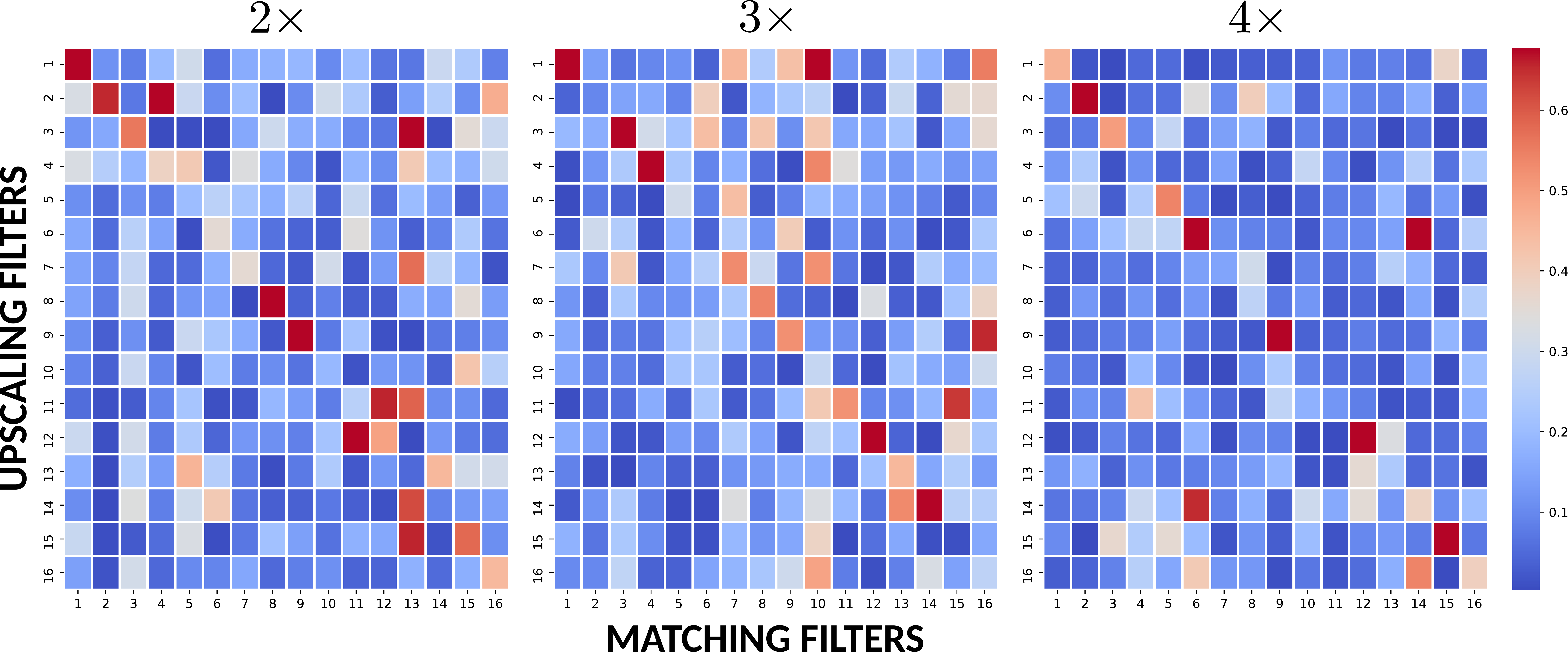}
    \caption{Correlations between \emph{upscaling} and \emph{matching} filters in eSR--TM $k=7, C=16$. Higher correlations along the diagonal mean that the model is performing template matching, with upscaling and matching filters that resemble a common template.}
    \label{fig:esr_correlation}
\end{figure}

The bold values in Table \ref{tab:imageq} highlight the best metrics for different columns, ignoring bicubic. edge--SR systems reach the best speed and lowest power consumption except for $4\times$ where ESPCN gets better. They also succeed to improve bicubic's image quality for small upscaling factors.

\begin{figure*}
  \begin{minipage}[b]{1.0\linewidth}
    \centering
    \centerline{\includegraphics[width=\linewidth]{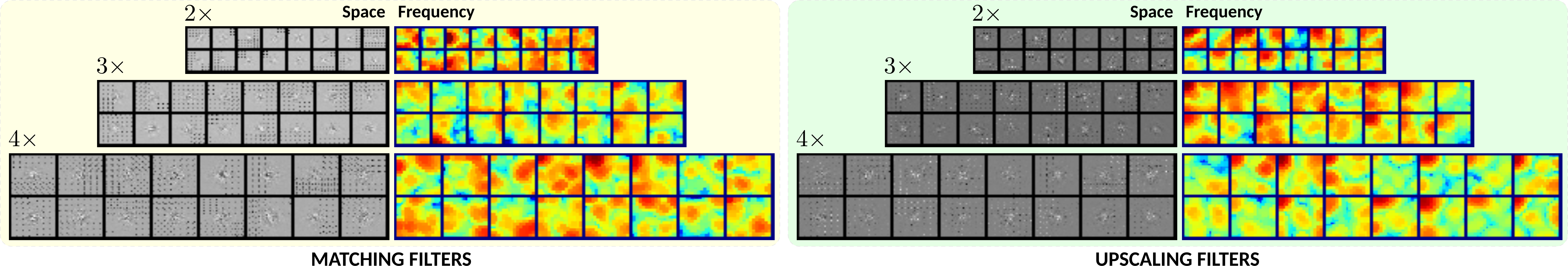}}
  \end{minipage}
\caption{Matching and upscaling filters obtained after training a one--layer architecture eSR--TM with kernel size $k=7$ and $C=18$ number of filters for $2\times$, $3\times$ and $4\times$ upscaling factors. Filters are displayed in the original spatial format as well as in frequency domain by using FFT visualization. The filters do not change smoothly within a single filter but show diverse directionality among different filters.}
\label{fig:esr_filters}
\end{figure*}

\begin{figure*}
  \begin{minipage}[b]{1.0\linewidth}
    \centering
    \centerline{\includegraphics[width=\linewidth]{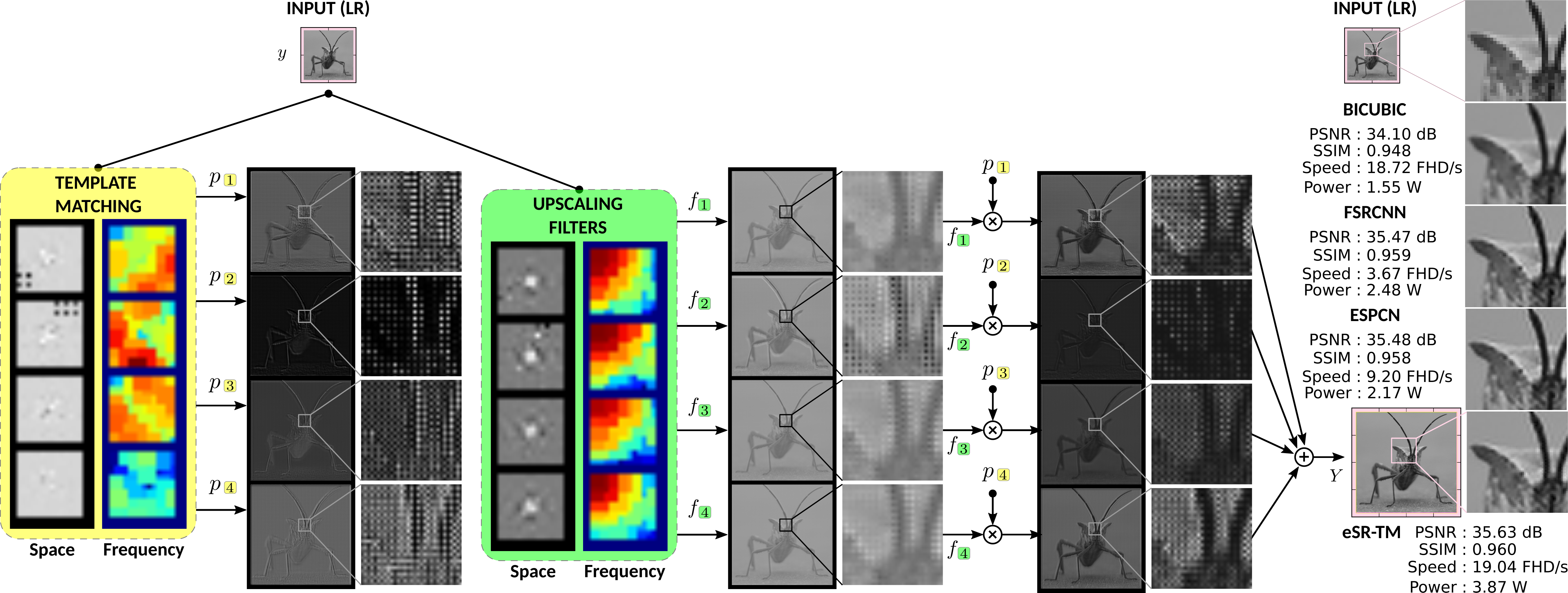}}
  \end{minipage}
\caption{Inspection of all intermediate outputs and filter coefficients for the eSR--TM $2\times$ architecture with kernel size $k=7$ and $C=4$ number of matching/upscaling filters. The diagram follows the interpretation in Figure \ref{fig:schematic}. Filters are displayed both in the original spatial format as well as in frequency domain by using FFT visualization. Each of the $4$ branches is focusing on a particular sub--pixel array.}
\label{fig:esr_interp}
\end{figure*}

\textbf{Filters.} In Figure \ref{fig:esr_interp} we display the step by step processing of $2\times$ upscaling using eSR--TM with kernel size $k=7$ and $C=4$ number of matching/upscaling filters. Here, we used equation \eqref{eq:shuffle_filters} to reconstruct the $4$ matching/upsampling filters from the efficient implementation containing $4\cdot 2^2=16$ filters. In addition to the filter coefficients we also display the FFT computed using a Kaiser--Bessel window for better frequency visualization~\cite{JGProakis_2007a}. The output for this particular image is about $1.5$ dB better than the bicubic output and it is displayed next to the outputs of ESPCN and FSRCNN models with similar image quality. Here, eSR--TM achieves roughly the same speed of bicubic upscaler.

The efficient filters use kernel size $k\times k$, and after multiplexing them with a pixel--shuffle layer we can recover the original filters of size $sk\times sk$. Thus, the filter sizes of eSR models grows with the upscaling factors as seen in Figure \ref{fig:esr_filters}. The filter coefficients in frequency domain show that each filter is processing different frequency bands. Although the filters are not smooth, they do show a level of discrimination between different directions.

Now, moving one step inside the network from the output in Figure \ref{fig:esr_interp}, we observe that the $4$ components of the sum are clearly focusing on different sub--pixel images. This pattern is also visible in the outputs of upscaling filters and template matching modules. Both matching and upscaling filters are not smooth and also show signs of different sub--pixel processing with some degree of directionality. This indicates that the different branches of the single convolutional layer used in eSR--TM are solving the upscaling problem independently for each sub--pixel image. This is in contrast with the smooth scaling filters used in the classical edge--directed interpolation\cite{VRAlgazi_1991a,XLi_2001a} and also compared to smooth directional filters observed in CNNs super--resolution interpretations in \cite{michelini2019tour}. Next, in Figure \ref{fig:esr_correlation} we compute the Pearson correlation between upscaling and matching filters for eSR--TM with $k=7$ and $C=16$.  The results show dominant correlations along the diagonal, stronger for $2\times$ factor and reducing strength towards $4\times$ factor. Strong correlations along the diagonal indicate a template matching strategy where upscaling and matching filters are similar for the same pattern and different to other patterns (see Figure \ref{fig:schematic}). Thus, we confirm that the training process has a tendency to converge towards a template matching strategy that is particularly strong for small upscaling factors.

\section{Conclusions}
The current trend in Edge--AI chips offers the chance to deploy efficient AI solutions at massive scale. But there is a vast range of performance requirements for which these solutions are unavailable for image SR. We propose the edge--SR architectures with the aim to fill the gap between classic and deep learning upscalers. We performed an exhaustive search among more than a thousand different models identically trained, revealing the gap between classic upscalers and deep--learning solutions. Our edge--SR configurations using a single convolutional layer showed promising results to fill this gap for small upscaling factors. The simplicity of the model also makes it interpretable and allows to visualize and understand all the intermediate steps of the process.

\section{Appendix}
\def\thesubsection{\Alph{subsection}}

\subsection{Evaluation Metrics}
\label{sec:metrics}

\begin{figure}
  \begin{minted}[
    frame=lines,
    framesep=2mm,
    baselinestretch=1.,
    bgcolor=shadecolor,
    fontsize=\scriptsize,
    linenos
  ]{python}
  import numpy as np
  import torch.autograd.profiler as profiler

  def str_to_time(s):
    if s.endswith('ms'):
      return float(s[:-2])*1e-3
    if s.endswith('us'):
      return float(s[:-2])*1e-6
    return float(s[:-1])

  def speed(model, input):
    dt = np.inf
    for _ in range(10):
        with profiler.profile(
          record_shapes=True, use_cuda=True
        ) as prof:
            with profiler.record_function(
              'model_inference'
            ):
                output = model(input)
        dt1 = str_to_time(
            prof.key_averages().table(
              sort_by='cpu_time_total', row_limit=10
            ).split(
              'CPU time total: '
            )[1].split('\n')[0]
        )
        dt2 = str_to_time(
            prof.key_averages().table(
              sort_by='cpu_time_total', row_limit=10
            ).split(
              'CUDA time total: '
            )[1][:-1]
        )
        dt = min(dt1+dt2, dt)

    pix = np.asarray(output.shape).prod()

    return pix / (dt * 1920*1080)

  \end{minted}
  \caption{Python function used to measure the speed of models in Pytorch v1.8. It runs a model $10$ times to avoid warm--up effects. Then, it parses the output of Pytorch profiler to get both CPU and GPU runtime. The speed is the number of output pixels divided by the minimum runtime in the $10$ runs.}
  \label{fig:speed_code}
\end{figure}
\textbf{Image quality.} Quantitative evaluations in our experiments include objective metrics PSNR and SSIM. These are reference--based metrics that measure the difference between an impaired image and ground truth. Higher values are better in both cases. The PSNR (range $0$ to $\infty$) is a log--scale version of mean--square--error and SSIM (range $0$ to $1$) uses image statistics to better correlate with human perception. Full expressions are as follows:
\begin{align}
    PSNR(X,Y) & = 10 \cdot \log_{10}\left(\frac{255^2}{MSE}\right) \;,\\
    SSIM(X,Y) & =\frac{(2\mu_X\mu_Y+c_1)(2\sigma_{XY}+c_2)}{(\mu_X^2+\mu_Y^2+c_1)(\sigma_X^2+\sigma_Y^2+c_2)} \;,
\end{align}
where $MSE=\mathbb{E}\left[(X-Y)^2\right]$ is the mean square error of the difference between $X$ and $Y$; $\mu_X$ and $\mu_Y$ are the averages of $X$ and $Y$, respectively; $\sigma_X^2$ and $\sigma_Y^2$ are the variances of $X$ and $Y$, respectively; $\sigma _{XY}$ is the covariance of X and Y; $c_1=6.5025$ and $c_2=58.5225$.
\label{sec:analysis}
\begin{figure}
  \centering
  \includegraphics[width=.8\linewidth]{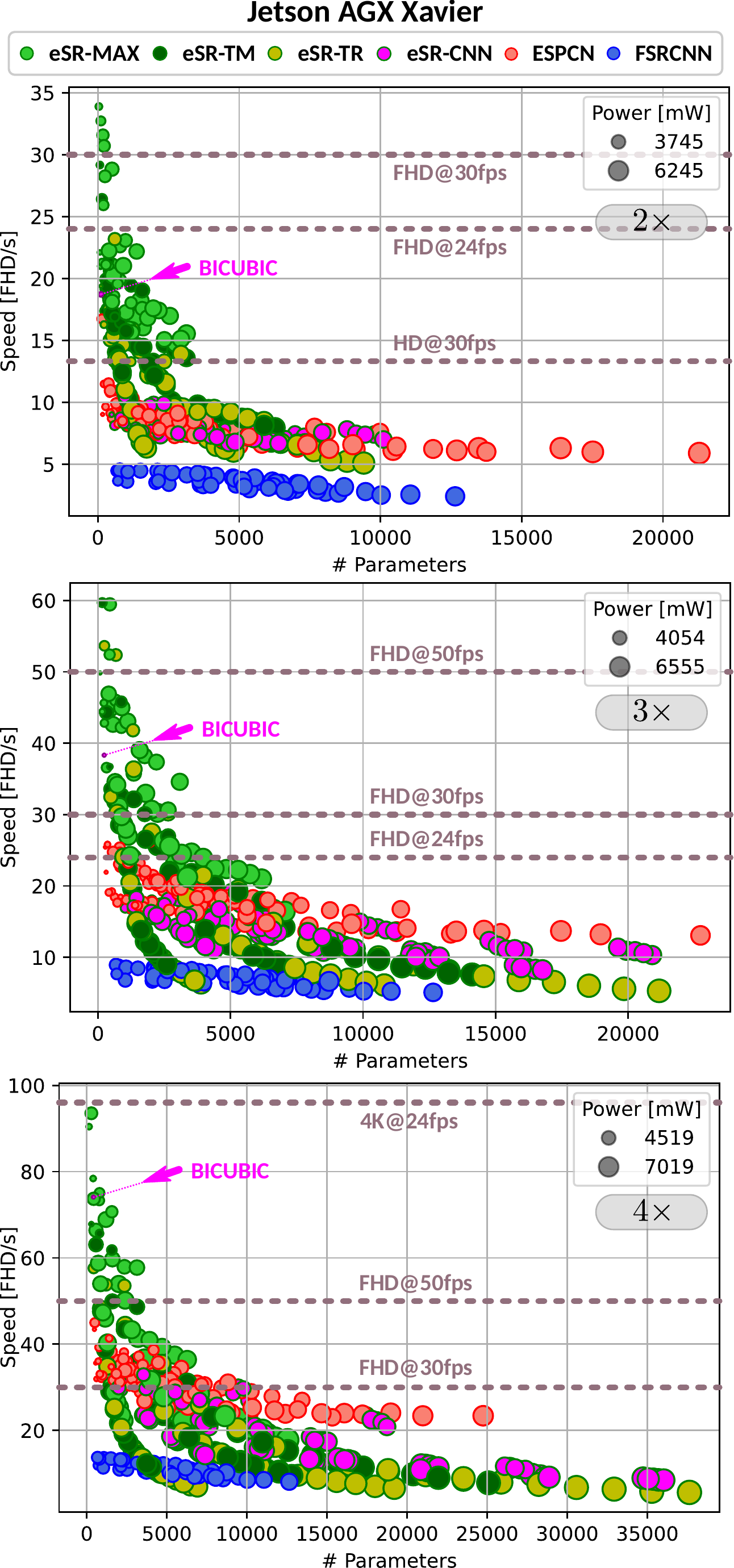}
  \caption{Example of the change in speed of the models with respect to their number of parameters for experiments on the Jetson AGX Xavier device. Models become faster at an exponential rate as the number of parameter reduces. \label{fig:parameters}}
\end{figure}

\begin{figure*}
  \centering
  \centerline{\includegraphics[width=\linewidth]{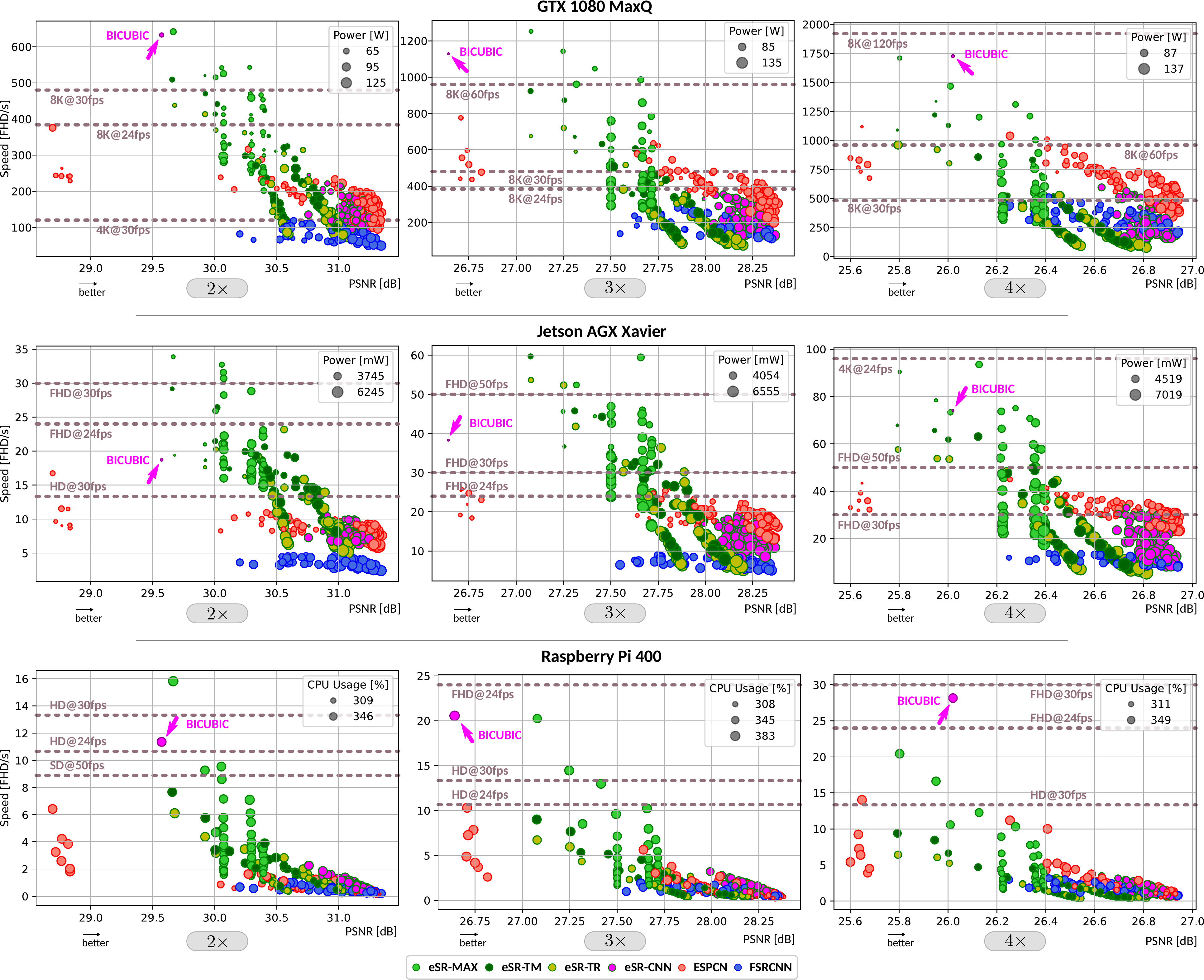}}
  \caption{Scatter plot to compare speed, in number of Full--HD pixels per second, with respect to quality, measured as PSNR for the BSDS--100 dataset. A total of $1,185$ models were identically trained considering different upscaling factors ($2\times$, $3\times$ and $4\times$) and architectures (eSR, ESPCN and FSRCNN). We run all models on edge devices: GTX--1080 Max--Q (GPU  with 16--bit floating point precision), Jetson AGX Xavier (GPU with 16--bit floating point precision) and Raspberry Pi 400 (CPU with 32--bit floating point precision). Magnified plots with model annotations are provided in the Figures \ref{fig:maxq_speed_vs_psnr_2x}, \ref{fig:maxq_speed_vs_psnr_3x}, \ref{fig:maxq_speed_vs_psnr_4x}, \ref{fig:agx_speed_vs_psnr_2x}, \ref{fig:agx_speed_vs_psnr_3x} and \ref{fig:agx_speed_vs_psnr_4x}.}
\label{fig:appendix-psnr_vs_speed}
\end{figure*}
\begin{figure*}
  \centering
  \centerline{\includegraphics[width=\linewidth]{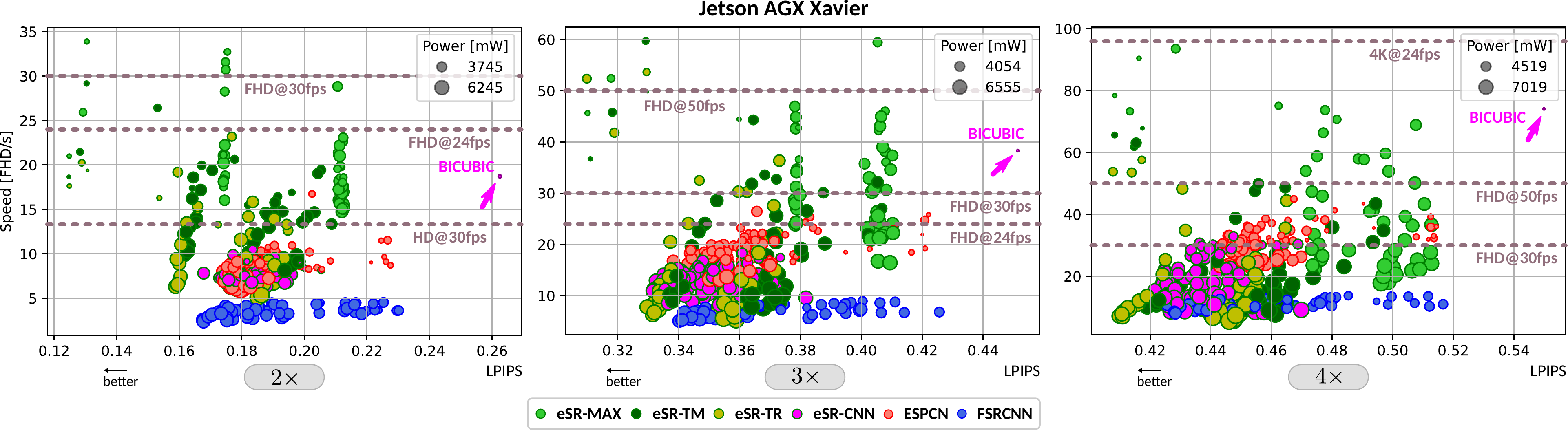}}
  \caption{Scatter plot to compare speed, in number of Full--HD pixels per second, with respect to quality, measured as LPIPS for the BSDS--100 dataset. Lower values of LPIPS mean better quality as opposed to PSNR where larger values are better. Compared with the same case but using PSNR quality measure in Figure \ref{fig:appendix-psnr_vs_speed}, eSR--TM performs much better and Bicubic shows the worst quality. }
\label{fig:appendix-lpips_vs_speed}
\end{figure*}

Benchmark in the literature can show big differences in PSNR and SSIM metrics due to different ways to evaluate color. For real--time applications it is common to process color images in YUV space, and the super--resolution task applies only to the luminance channel $Y$. Color in $U$ and $V$ channels can use a faster bicubic upscaler with small impact in perceptual quality. We follow the implementation in \cite{psnr-ssim-y} using a conversion of RGB to YUV color--spaces following the BT.709 standard, including offsets that are often avoided in other implementations.

In Appendix \ref{sec:analysis} and \ref{sec:reproducibility} we also provide measurements of the perceptual quality metric LPIPS defined in \cite{zhang2018unreasonable} and implemented using the PIQA library \cite{piqa}.

\textbf{Speed.} We run each model to output a set of $14$ Full--HD images, downscaling appropriately from randomly selected images of the DIV2K dataset~\cite{Agustsson_2017_CVPR_Workshops}. We use 16--bit floating point precision during inference. For each image we run the model $10$ times to avoid warm--up effects, measuring: the minimum CPU and GPU processing time from profiler's data. We computed the speed of a model using the total number of pixels processed (considering only one run per image) divided by the processing time (using the minimum time over each one of the $10$ runs). To make the measurement of speed easier to read we use units of \texttt{[FHD/s]}, this is, number of Full--HD pixels ($1920\times 1080$) per second. Figure \ref{fig:speed_code} shows the code used to parse Pytorch's profiler output to obtain the CPU and GPU processing time.

\textbf{Power and CPU usage.} We run each model to output a set of $14$ Full--HD images, downscaling appropriately from randomly selected images of the DIV2K dataset~\cite{Agustsson_2017_CVPR_Workshops}. We use 16--bit floating point precision during inference. During this process we monitor the maximum power consumption using \texttt{nvidia-smi} for GTX 1080 Max--Q GPU, and \texttt{tegrastats} for Jetson AGX Xavier. We register the maximum power measured in this process. The power data provided by Max--Q driver is in units of watts, whereas the AGX device uses units of milliwatts. The AGX device allows different profiles for power consumption and in our experiments we used 30 Watts.

The Raspberry Pi 400 device does not include power sensors and in this case we replace the power measurement by CPU usage, monitor by parsing the \texttt{top} command with delay--time of $0.05s$ and registering the average CPU reading during the inference process.

\begin{figure}
  \begin{minted}[
    frame=lines,
    framesep=2mm,
    baselinestretch=1.,
    bgcolor=shadecolor,
    fontsize=\scriptsize,
    linenos
  ]{python}
>>> import pickle
>>> test = pickle.load(open('tests.pkl', 'rb'))
>>> test['Bicubic_s2']
  {'psnr_Set5': 33.72849620514912,
   'ssim_Set5': 0.9283912810369976,
   'lpips_Set5': 0.14221979230642318,
   'psnr_Set14': 30.286027790636204,
   'ssim_Set14': 0.8694934108301432,
   'lpips_Set14': 0.19383049915943826,
   'psnr_BSDS100': 29.571233006609656,
   'ssim_BSDS100': 0.8418117904964167,
   'lpips_BSDS100': 0.26246454380452633,
   'psnr_Urban100': 26.89378248655882,
   'ssim_Urban100': 0.8407461069831571,
   'lpips_Urban100': 0.21186692919582129,
   'psnr_Manga109': 30.850672809780587,
   'ssim_Manga109': 0.9340133711400112,
   'lpips_Manga109': 0.102985977955641,
   'parameters': 104,
   'speed_AGX': 18.72132628065749,
   'power_AGX': 1550,
   'speed_MaxQ': 632.5429857814075,
   'power_MaxQ': 50,
   'temperature_MaxQ': 76,
   'memory_MaxQ': 2961,
   'speed_RPI': 11.361346064182795,
   'usage_RPI': 372.8714285714285}

  \end{minted}
  \caption{Example of how to read our test data from the Python dictionary file \href{https://www.dropbox.com/s/c52or2cyqfofpxg/tests.pkl}{\texttt{tests.pkl}}.}
  \label{fig:tests}
\end{figure}

\begin{figure}
  \begin{minted}[
    frame=lines,
    framesep=1mm,
    baselinestretch=1.,
    bgcolor=shadecolor,
    fontsize=\scriptsize,
    linenos
  ]{python}
  import torch
  from torch import nn

  class edgeSR_MAX(nn.Module):
      def __init__(self, C, k, s):
          super().__init__()

          self.pixel_shuffle = nn.PixelShuffle(s)
          self.filter = nn.Conv2d(
              in_channels=1,
              out_channels=s*s*C,
              kernel_size=k,
              stride=1,
              padding=(k-1)//2,
              bias=False,
          )

      def forward(self, input):
          return self.pixel_shuffle(
              self.filter(input)
          ).max(dim=1, keepdim=True)[0]

  \end{minted}
  \caption{Pytorch v1.8 implementation of edge--SR Maximum (eSR--MAX) single--layer architecture.}
  \label{fig:esrmax}
\end{figure}

\begin{figure}
  \begin{minted}[
    frame=lines,
    framesep=1mm,
    baselinestretch=1.,
    bgcolor=shadecolor,
    fontsize=\scriptsize,
    linenos
  ]{python}
  import torch
  from torch import nn

  class edgeSR_TM(nn.Module):
      def __init__(self, C, k, s):
          super().__init__()

          self.pixel_shuffle = nn.PixelShuffle(s)
          self.softmax = nn.Softmax(dim=1)
          self.filter = nn.Conv2d(
              in_channels=1,
              out_channels=2*s*s*C,
              kernel_size=k,
              stride=1,
              padding=(k-1)//2,
              bias=False,
          )

      def forward(self, input):
          filtered = self.pixel_shuffle(
              self.filter(input)
          )
          B, C, H, W = filtered.shape

          filtered = filtered.view(B, 2, C, H, W)
          upscaling = filtered[:, 0]
          matching = filtered[:, 1]
          return torch.sum(
              upscaling * self.softmax(matching),
              dim=1, keepdim=True
          )

  \end{minted}
  \caption{Pytorch v1.8 implementation of edge--SR Template Matching (eSR--TM) single--layer architecture.}
  \label{fig:esrtm}
\end{figure}

\begin{figure}
  \begin{minted}[
    frame=lines,
    framesep=1mm,
    baselinestretch=1.,
    bgcolor=shadecolor,
    fontsize=\scriptsize,
    linenos
  ]{python}
  import torch
  from torch import nn

  class edgeSR_TR(nn.Module):
      def __init__(self, C, k, s):
          super().__init__()

          self.pixel_shuffle = nn.PixelShuffle(s)
          self.softmax = nn.Softmax(dim=1)
          self.filter = nn.Conv2d(
              in_channels=1,
              out_channels=3*s*s*C,
              kernel_size=k,
              stride=1,
              padding=(k-1)//2,
              bias=False,
          )

      def forward(self, input):
          filtered = self.pixel_shuffle(
              self.filter(input)
          )
          B, C, H, W = filtered.shape

          filtered = filtered.view(B, 3, C, H, W)
          value = filtered[:, 0]
          query = filtered[:, 1]
          key = filtered[:, 2]
          return torch.sum(
              value * self.softmax(query*key),
              dim=1, keepdim=True
          )

  \end{minted}
  \caption{Pytorch v1.8 implementation of edge--SR TRansformer (eSR--TR) single--layer architecture.}
  \label{fig:esrtr}
\end{figure}

\begin{figure}
  \begin{minted}[
    frame=lines,
    framesep=1mm,
    baselinestretch=1.,
    bgcolor=shadecolor,
    fontsize=\scriptsize,
    linenos
  ]{python}
  import torch
  from torch import nn

  class edgeSR_CNN(nn.Module):
      def __init__(self, C, D, S, s):
          super().__init__()

          self.softmax = nn.Softmax(dim=1)
          if D == 0:
              self.filter = nn.Sequential(
                  nn.Conv2d(D, S, 3, 1, 1),
                  nn.Tanh(),
                  nn.Conv2d(
                      in_channels=S,
                      out_channels=2*s*s*C,
                      kernel_size=3,
                      stride=1,
                      padding=1,
                      bias=False,
                  ),
                  nn.PixelShuffle(s),
              )
          else:
              self.filter = nn.Sequential(
                  nn.Conv2d(1, D, 5, 1, 2),
                  nn.Tanh(),
                  nn.Conv2d(D, S, 3, 1, 1),
                  nn.Tanh(),
                  nn.Conv2d(
                      in_channels=S,
                      out_channels=2*s*s*C,
                      kernel_size=3,
                      stride=1,
                      padding=1,
                      bias=False,
                  ),
                  nn.PixelShuffle(s),
              )

      def forward(self, input):
          filtered = self.filter(input)
          B, C, H, W = filtered.shape

          filtered = filtered.view(B, 2, C, H, W)
          upscaling = filtered[:, 0]
          matching = filtered[:, 1]
          return torch.sum(
              upscaling * self.softmax(matching),
              dim=1, keepdim=True
          )

  \end{minted}
  \caption{Pytorch v1.8 implementation of edge--SR CNN (eSR--CNN) multi--layer architecture.}
  \label{fig:esrcnn}
\end{figure}

\subsection{Extended Analysis}
\textbf{Speed vs parameters.} In Figure \ref{fig:parameters} we show the relationship between the size of the model, given by the number of parameters, and the execution speed when running the models on GPU devices. We observe that smaller models run faster, and the speed increases exponentially. The non--linear relation between speed and number of parameters becomes critical under: $5,000$ parameters for $2\times$ upscaling factor, $10,000$  parameters for $3\times$ upscaling factor, and $15,000$ parameters for $4\times$ upscaling factor. This shows the importance of focusing on speed compared to number of parameters in our study. Research on lightweight SR architectures often focuses on number of parameters and typically use several hundred thousand parameters where the non--linearity is still not critical.

\textbf{Additional device.} In Figure \ref{fig:appendix-psnr_vs_speed} we show scatter plots including Nvidia GeForce GTX--1080 Max--Q GPU. This is a mobile high--end GPU from the Pascal series typically used for laptop computers. The power consumption of the Max--Q design ranges between 90 and 110 Watt, compared to 30 Watt used in the Jetson AGX Xavier. Although much more powerful than a Raspberry Pi and Jetson AGX devices, the GTX 1080 Max--Q device can fit in high--end display TV panels and thus serve a different range of applications for edge devices. Consequently, Figure \ref{fig:appendix-psnr_vs_speed} shows a performance that can deliver 8K videos in real--time (even with 16--bit floating point precision). We also observe that ESPCN performance improves for $3\times$ and $4\times$ upscaling factors, indicating the important effect of the increased number of cores in GPU architectures as well as a significant increase in power consumption.

\begin{figure*}
  \centering
  \includegraphics[width=\linewidth]{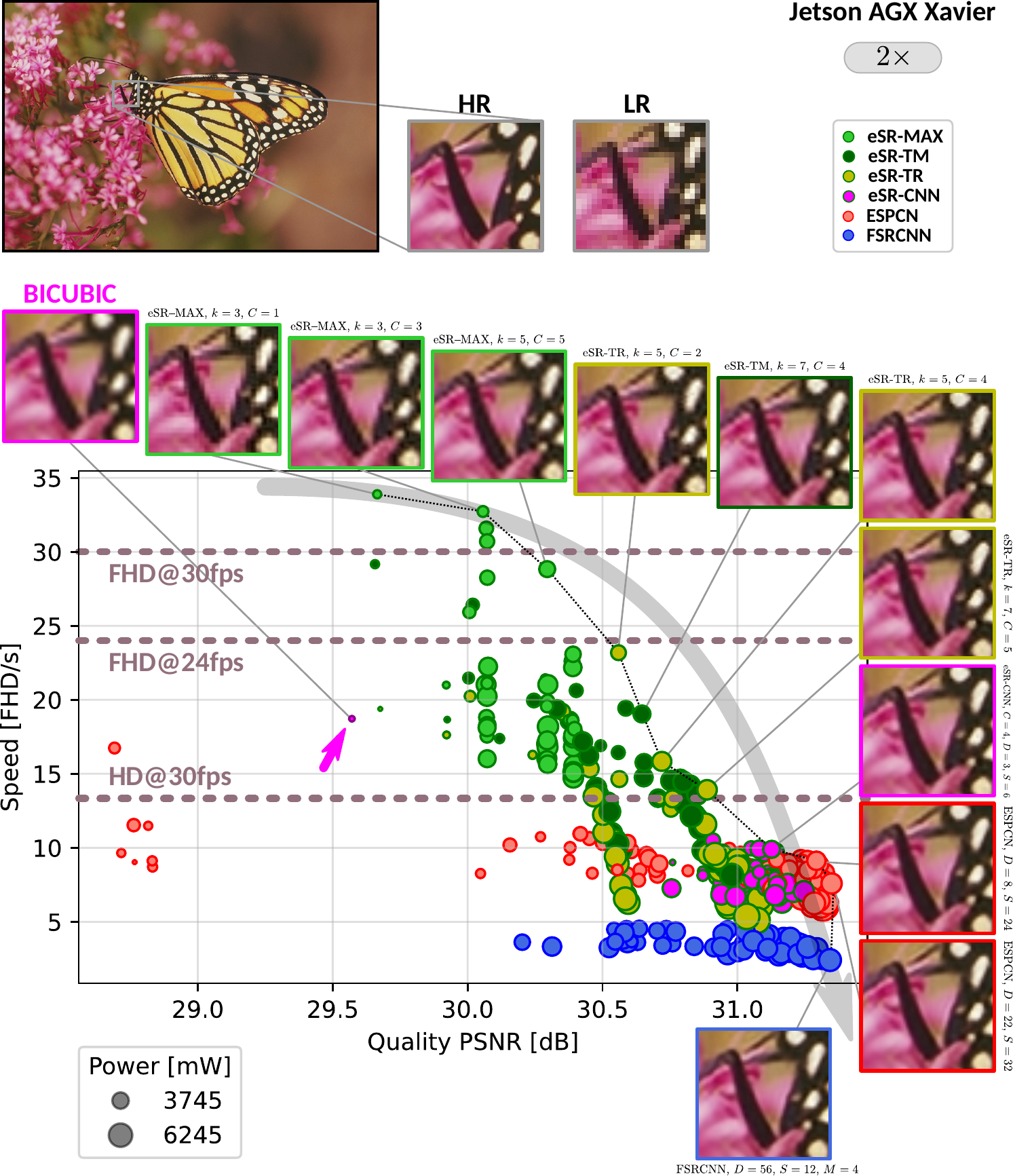}
  \caption{Example output images in the trajectory through the trade--off between image quality and runtime performance for $2\times$ upscaling on a Jetson AGX Xavier edge device. \label{fig:qualitative_2x}}
\end{figure*}

\begin{figure*}
  \centering
  \includegraphics[width=\linewidth]{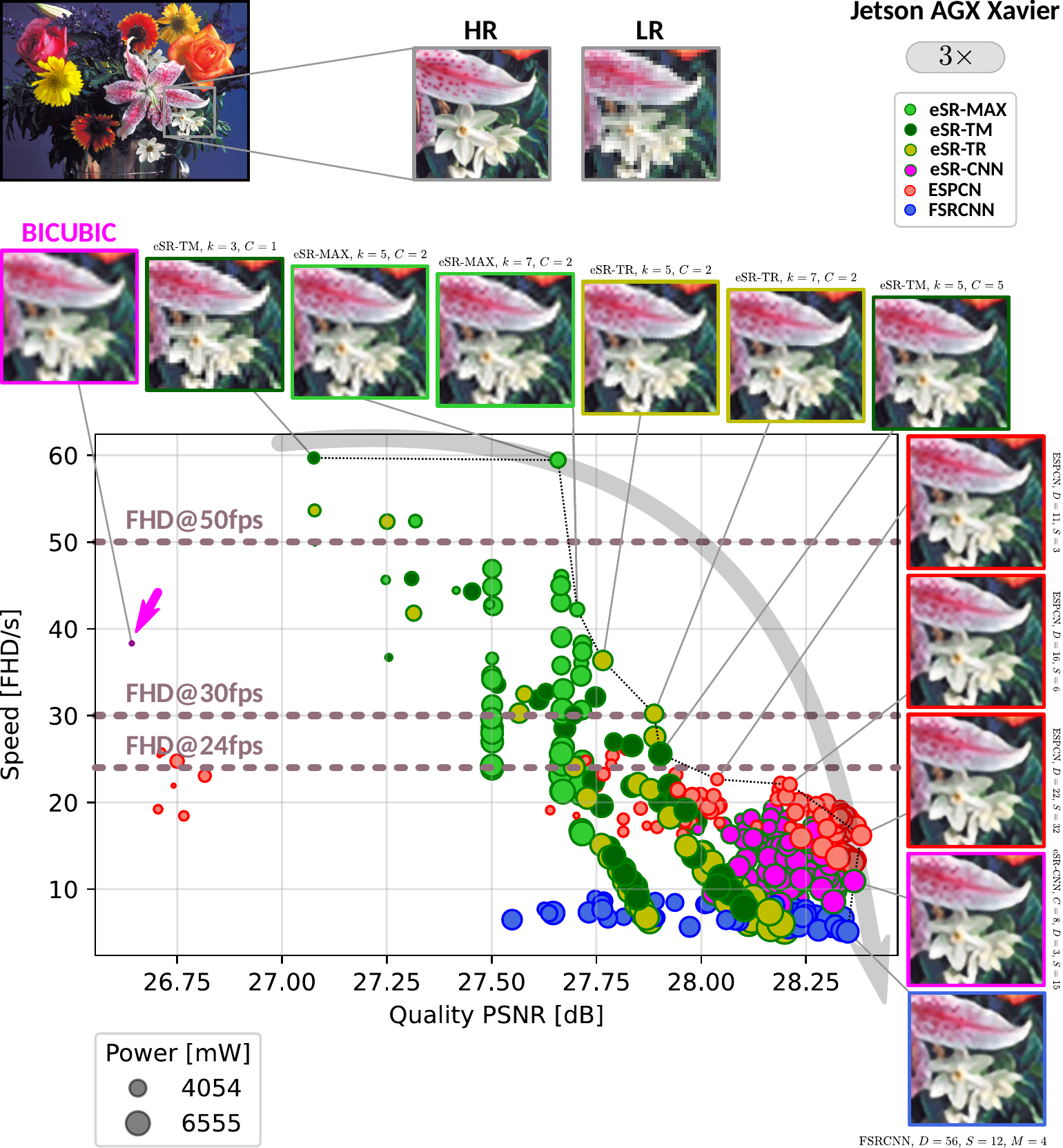}
  \caption{Example output images in the trajectory through the trade--off between image quality and runtime performance for $3\times$ upscaling on a Jetson AGX Xavier edge device. \label{fig:qualitative_3x}}
\end{figure*}

\begin{figure*}
  \centering
  \includegraphics[width=\linewidth]{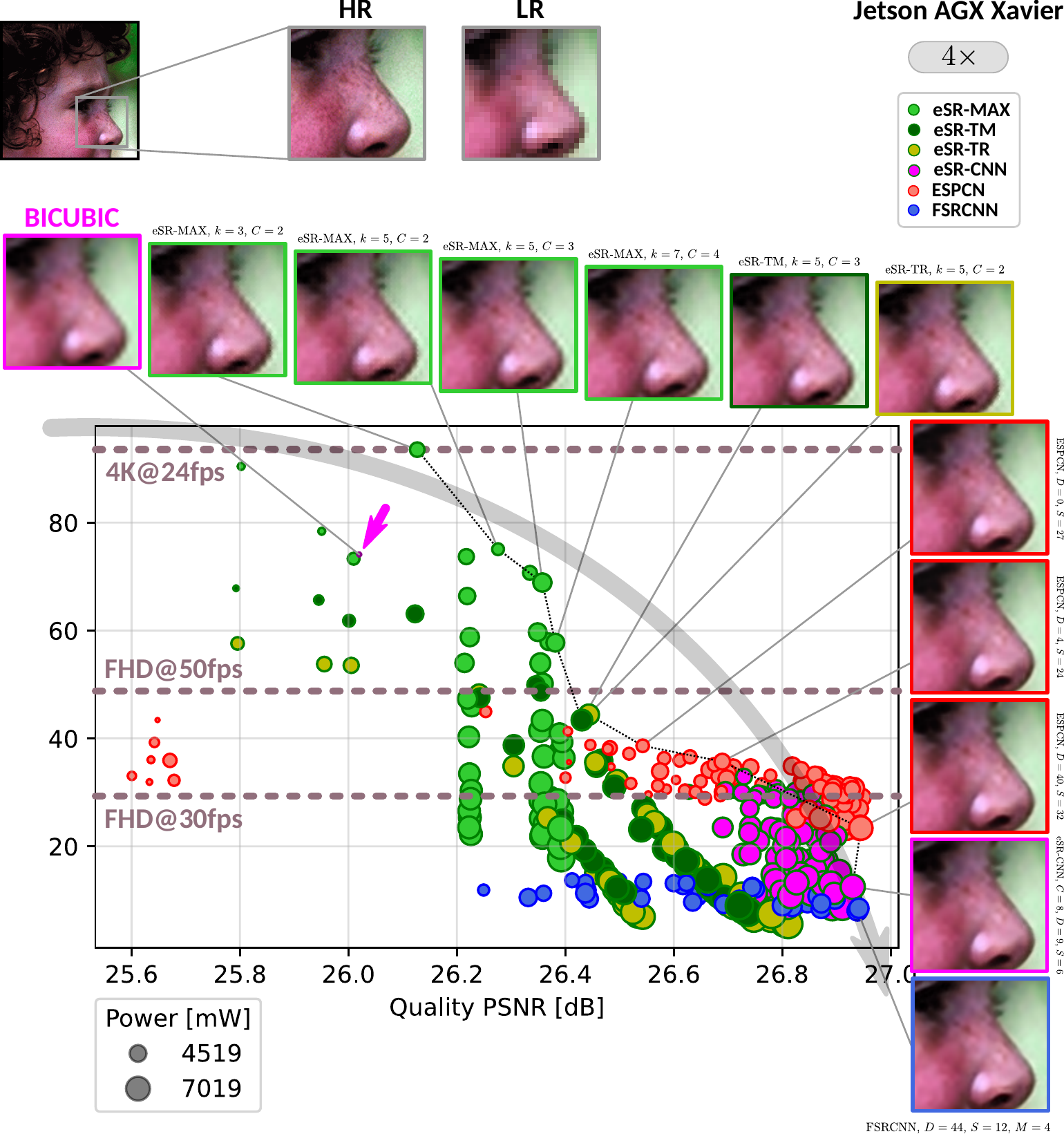}
  \caption{Example output images in the trajectory through the trade--off between image quality and runtime performance for $4\times$ upscaling on a Jetson AGX Xavier edge device. \label{fig:qualitative_4x}}
\end{figure*}

\textbf{Qualitative evaluation.} In Figures \ref{fig:qualitative_2x}, \ref{fig:qualitative_3x} and \ref{fig:qualitative_4x} we show example output images for different models. These models were selected by moving in an approximately optimal trajectory in the Speed vs Quality (SQ) plane for the Jetson AGX Xavier device. As mentioned before, the best advantage of edge--SR models is observed for $2\times$ upscaling factor. This is both the most difficult and important factor for applications due to the high input throughput in high resolution displays (e.g. HD to FHD). The classic bicubic upscalers offers over--smooth outputs that are somehow effective to reduce \emph{jaggy} artifacts. edge--MAX models can significantly improve sharpness with limited control over jaggies. Next, edge--TM and edge--TR models show the best trade--off with very similar performance. These models are more effective at reducing jaggies before multi--layer networks like edge--CNN and ESPCN become better with both sharp and smooth edges.

In Figure \ref{fig:appendix-lpips_vs_speed} we show a few scatter plots of the trade--off between image quality and runtime performance using the perceptual quality metric LPIPS~\cite{zhang2018unreasonable} for the BSDS--100 dataset. Here, lower values of LPIPS mean better quality as opposed to PSNR where larger values are better. Compared with the same case but using PSNR quality measure in Figure \ref{fig:appendix-psnr_vs_speed}, eSR--TM performs much better and Bicubic shows the worst quality.

At $3\times$ and $4\times$ upscaling factor the pattern is similar but it becomes more difficult for single--layer models to effectively reduce jaggy artifacts. Results get worsen and we observe an increased gap between bicubic and other architectures. ESPCN becomes better at high quality ranges and overcomes edge--SR models for the most part. Finally we note that trade--off evaluations at $3\times$ and $4\times$ upscaling factors could be misleading, as flickering video artifacts are likely to become visible at this point and video SR solutions might be needed. For this reason, the results at $2\times$ are arguably the most important for practical applications.

\begin{figure*}
  \centering\vspace*{-.6cm}
  \includegraphics[height=\textheight]{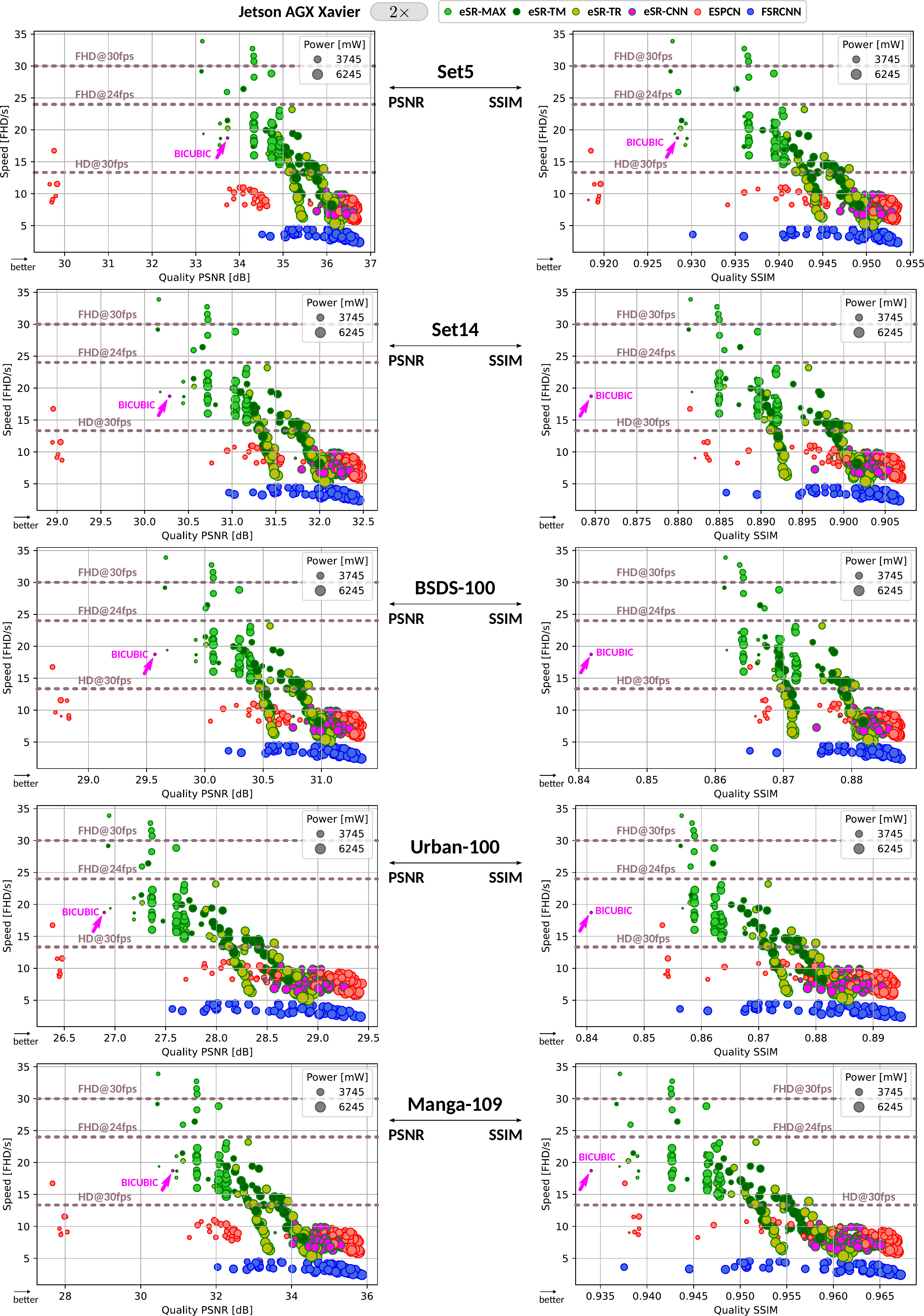}
  \caption{Example of the effect of different metric (PSNR/ SSIM) and datasets in the trade--off between image quality and runtime performance. We observe that even though the range of values and relative positions change, edge--SR models remain with better performance at high speed, and multi--layer networks remain better at low speed. \label{fig:ssim_and_datasets}}
\end{figure*}

\textbf{Effect of datasets and metrics.} Scatter plots in the main text use only PSNR metric measured in the BSDS--100 dataset. In Figure \ref{fig:ssim_and_datasets} we show the effect of changing both the metric, from PSNR to SSIM, and dataset, among Set5, Set14, BSDS--100, Urban--100 and Manga--109. The range of values and relative position of scatter points changes. For example, SSIM shows a larger margin in image quality between bicubic and other models. PSNR values show significant changes depending on the dataset. Nevertheless, the trade--off trajectory and the advantage shown by different architectures remains the same.

\begin{figure*}
  \centering\vspace*{-.5cm}\hspace*{-1.7cm}
  \includegraphics[width=1.22\textwidth]{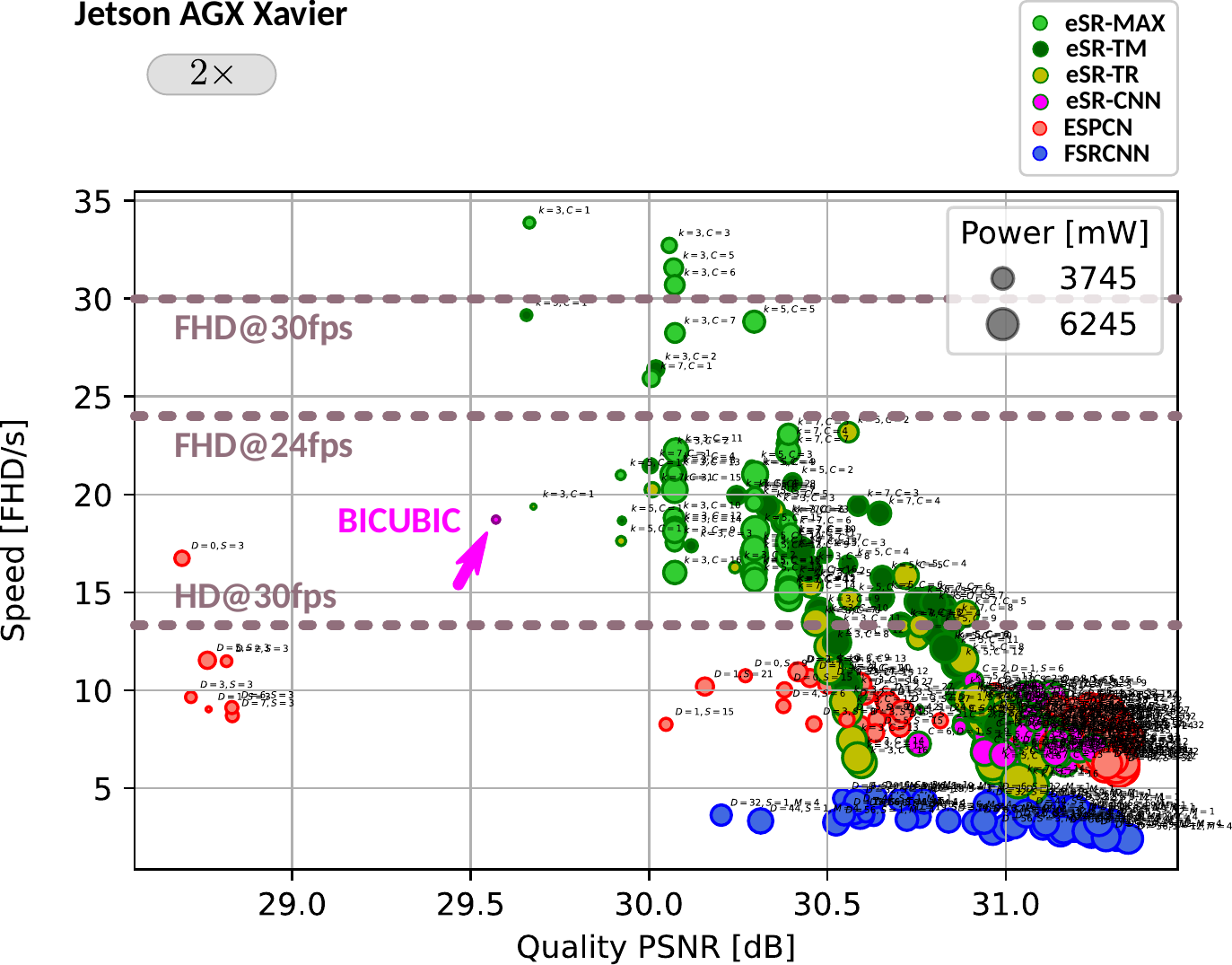}
  \caption{Speed of $2\times$ image SR models, measured in Full--HD pixels per second on a Jetson AGX Xavier GPU using 16--bit floating point precision, with respect to image quality, measure as PSNR in the BSDS--100 dataset. \label{fig:agx_speed_vs_psnr_2x}}
\end{figure*}

\begin{figure*}
  \centering\vspace*{-.5cm}\hspace*{-1.7cm}
  \includegraphics[width=1.22\textwidth]{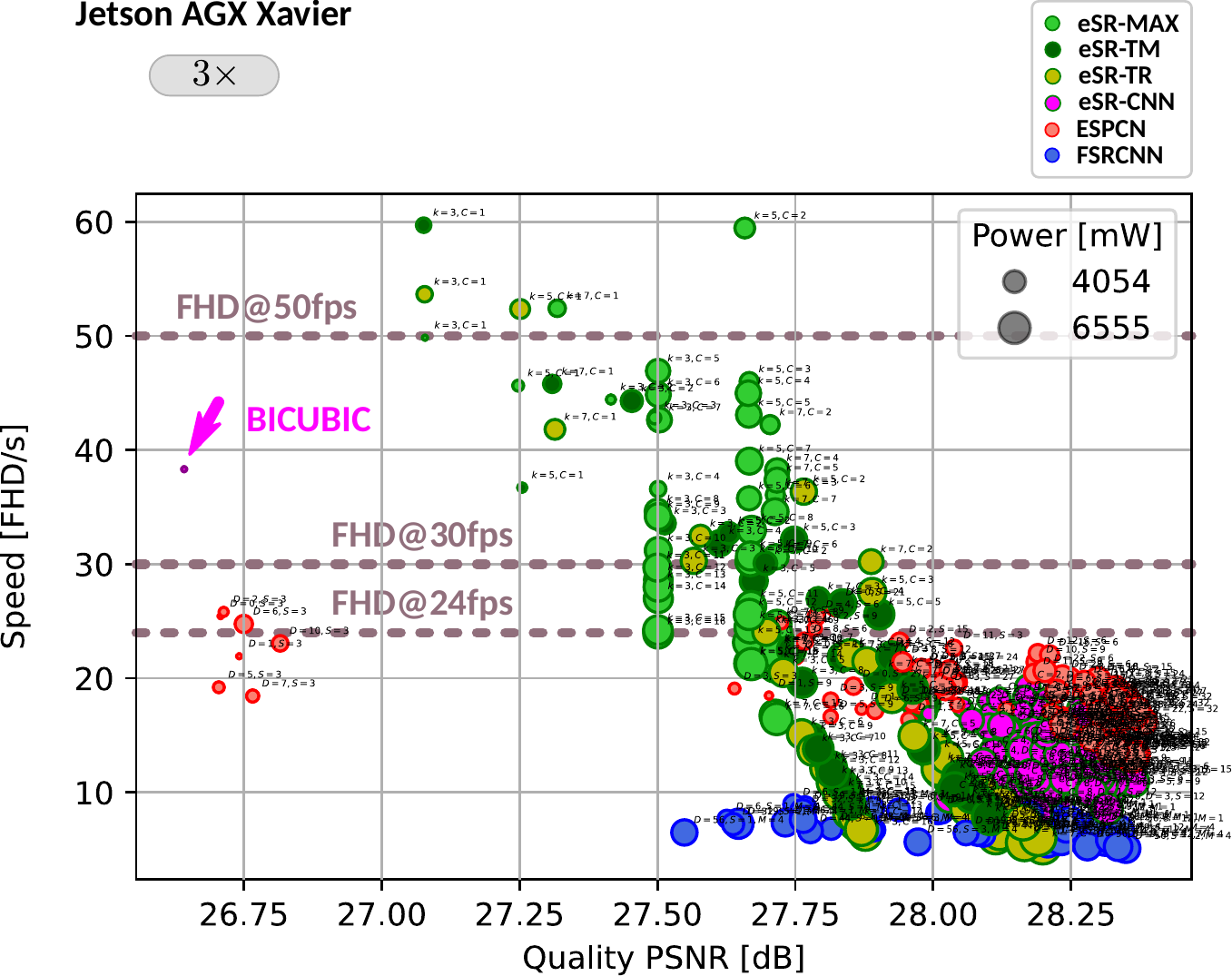}
  \caption{Speed of $3\times$ image SR models, measured in Full--HD pixels per second on a Jetson AGX Xavier GPU using 16--bit floating point precision, with respect to image quality, measure as PSNR in the BSDS--100 dataset. \label{fig:agx_speed_vs_psnr_3x}}
\end{figure*}

\begin{figure*}
  \centering\vspace*{-.5cm}\hspace*{-1.7cm}
  \includegraphics[width=1.22\textwidth]{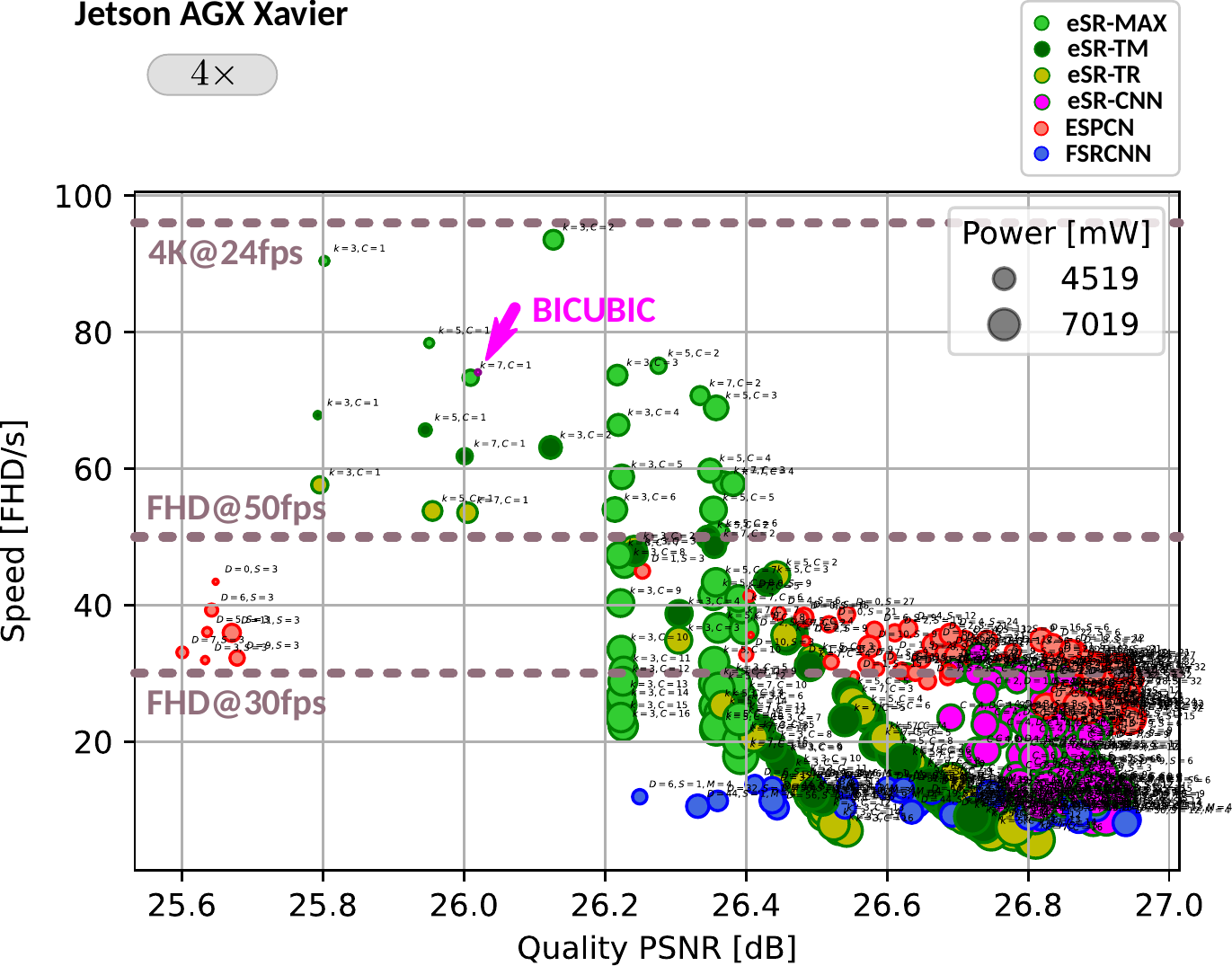}
  \caption{Speed of $4\times$ image SR models, measured in Full--HD pixels per second on a Jetson AGX Xavier GPU using 16--bit floating point precision, with respect to image quality, measure as PSNR in the BSDS--100 dataset. \label{fig:agx_speed_vs_psnr_4x}}
\end{figure*}

\begin{figure*}
  \centering\vspace*{-.5cm}\hspace*{-1.7cm}
  \includegraphics[width=1.22\textwidth]{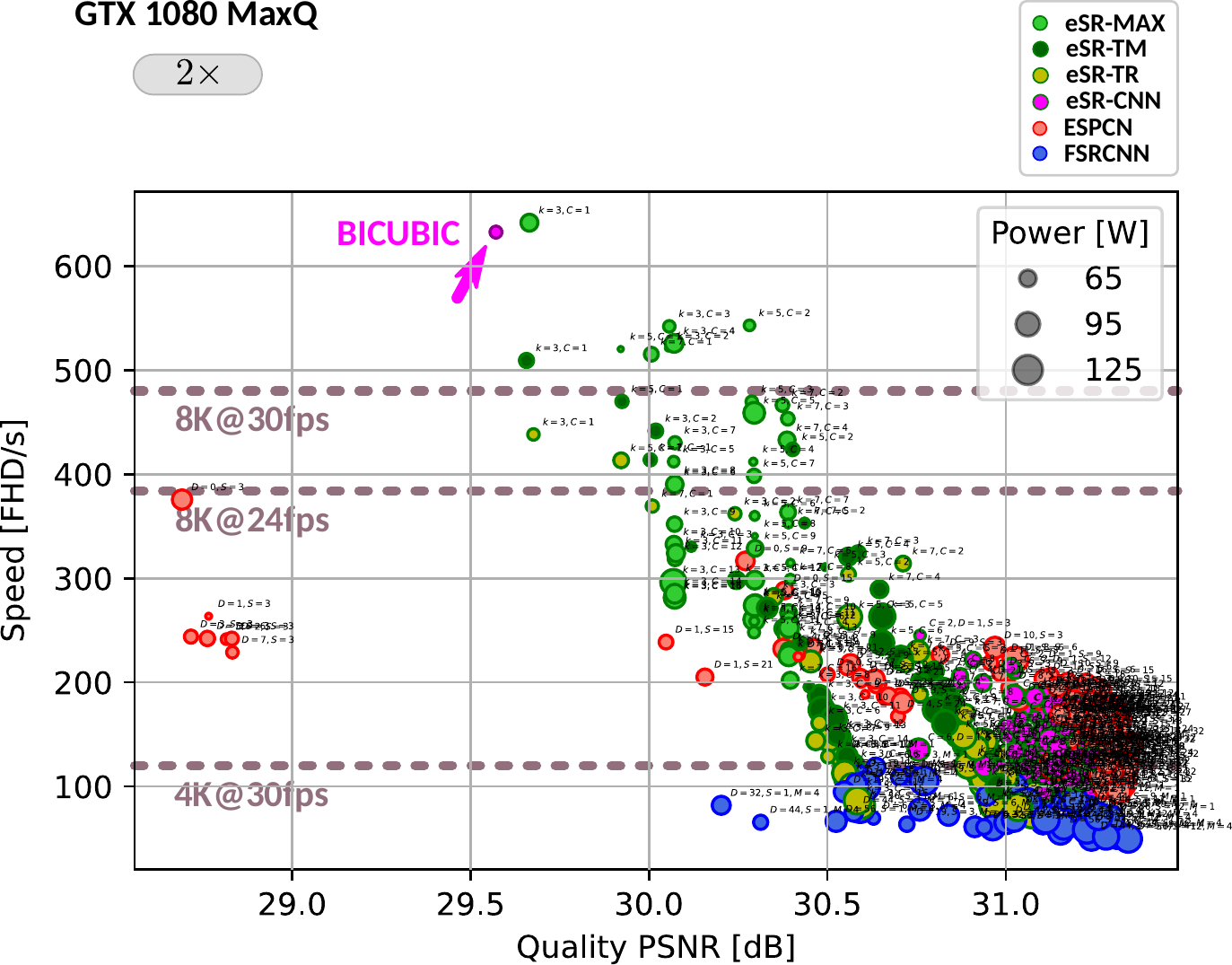}
  \caption{Speed of $2\times$ image SR models, measured in Full--HD pixels per second on a GTX 1080 Max--Q GPU using 16--bit floating point precision, with respect to image quality, measure as PSNR in the BSDS--100 dataset. \label{fig:maxq_speed_vs_psnr_2x}}
\end{figure*}

\begin{figure*}
  \centering\vspace*{-.5cm}\hspace*{-1.7cm}
  \includegraphics[width=1.22\textwidth]{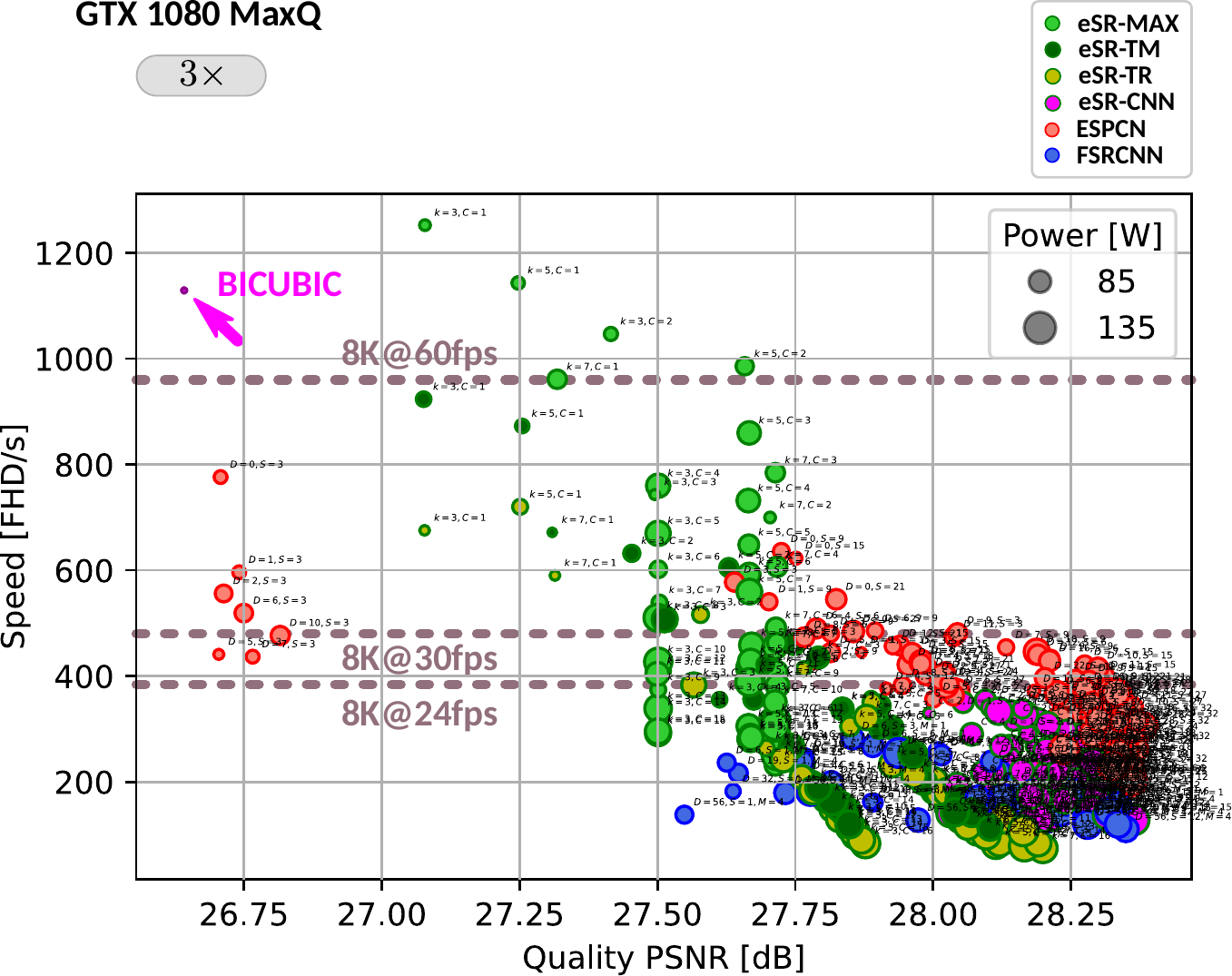}
  \caption{Speed of $3\times$ image SR models, measured in Full--HD pixels per second on a GTX 1080 Max--Q GPU using 16--bit floating point precision, with respect to image quality, measure as PSNR in the BSDS--100 dataset. \label{fig:maxq_speed_vs_psnr_3x}}
\end{figure*}

\begin{figure*}
  \centering\vspace*{-.5cm}\hspace*{-1.7cm}
  \includegraphics[width=1.22\textwidth]{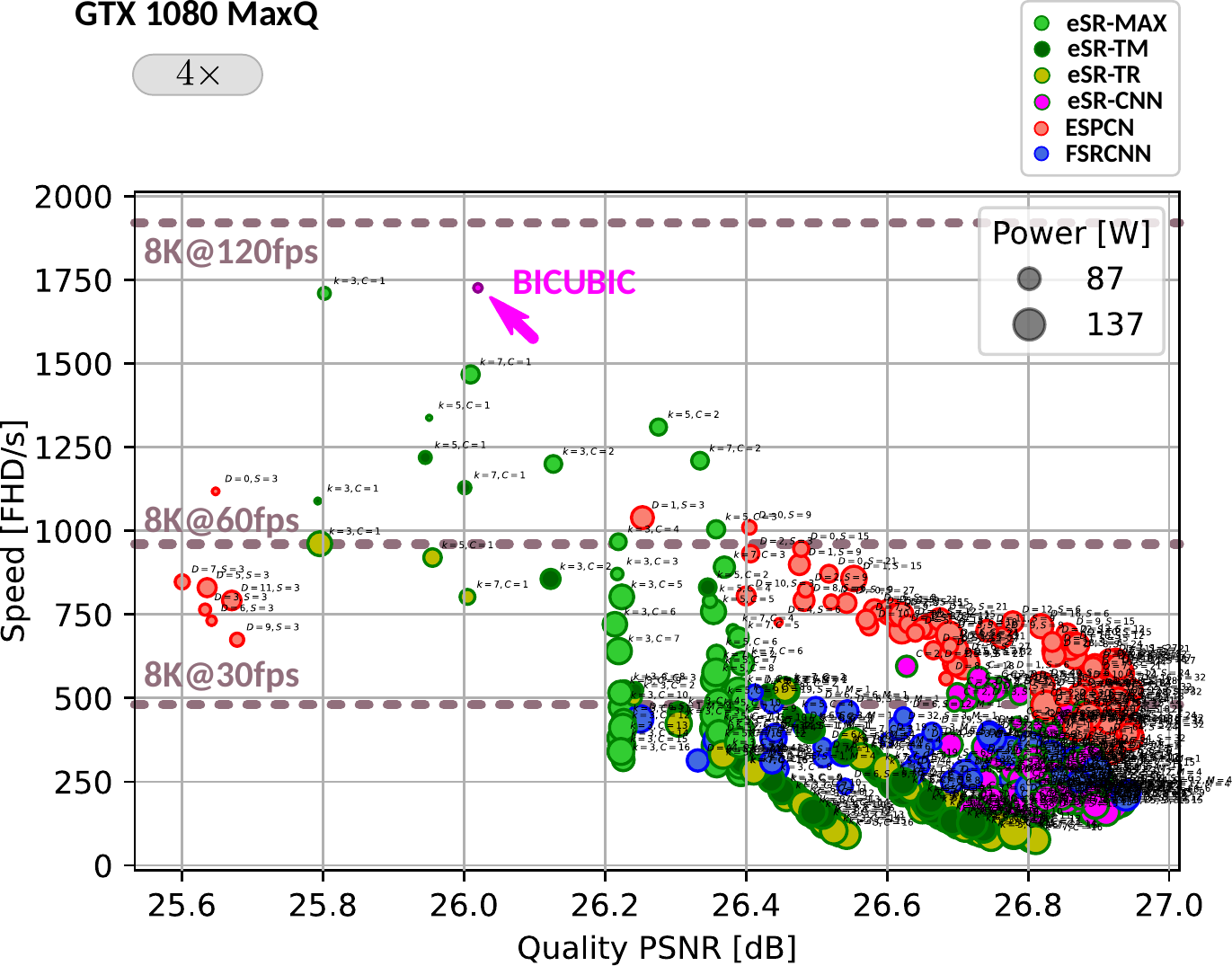}
  \caption{Speed of $4\times$ image SR models, measured in Full--HD pixels per second on a GTX 1080 Max--Q GPU using 16--bit floating point precision, with respect to image quality, measure as PSNR in the BSDS--100 dataset. \label{fig:maxq_speed_vs_psnr_4x}}
\end{figure*}

\textbf{Magnified scatter plots.} It is useful to identify the hyper--parameters of each specific model in scatter plots. Figures \ref{fig:maxq_speed_vs_psnr_2x}, \ref{fig:maxq_speed_vs_psnr_3x}, \ref{fig:maxq_speed_vs_psnr_4x}, \ref{fig:agx_speed_vs_psnr_2x}, \ref{fig:agx_speed_vs_psnr_3x} and \ref{fig:agx_speed_vs_psnr_4x} show magnified plots using all the page width and include annotations with model hyper--parameters. The annotations are useful at middle and high speed ranges where data is more sparse. In the high image quality range the performance of different models become clustered and the annotations are not readable. Here, we recommend to look at Figures \ref{fig:qualitative_2x}, \ref{fig:qualitative_3x} and \ref{fig:qualitative_4x} to identify the best models. Finally, we make all test data available in a Python dictionary file (see Appendix \ref{sec:reproducibility}).

\subsection{Reproducibility}
\label{sec:reproducibility}

\textbf{Test results.} Test results are provided in the Python dictionary file \href{https://www.dropbox.com/s/c52or2cyqfofpxg/tests.pkl}{\texttt{tests.pkl}} using the native \emph{pickle} module. A sample code to read the file is provided in Figure \ref{fig:tests}. The keys of the dictionary identify the name of each model and its hyper--parameters using the following format:
\begin{itemize}
    \setlength\itemsep{0em}
    \item \texttt{'Bicubic\_s\#'},
    \item \texttt{'eSR-MAX\_s\#\_K\#\_C\#'},
    \item \texttt{'eSR-TM\_s\#\_K\#\_C\#'},
    \item \texttt{'eSR-TR\_s\#\_K\#\_C\#'},
    \item \texttt{'eSR-CNN\_s\#\_C\#\_D\#\_S\#'},
    \item \texttt{'ESPCN\_s\#\_D\#\_S\#'}, or
    \item \texttt{'FSRCNN\_s\#\_D\#\_S\#\_M\#'},
\end{itemize}
where \texttt{\#} represents an integer number with the value of the correspondent hyper--parameter. For each model the data of the dictionary contains a second dictionary with the information displayed in Figure \ref{fig:tests}. This includes: number of model parameters; image quality metrics PSNR, SSIM and LPIPS measured in $5$ different datasets; as well as power, speed, CPU usage, temperature and memory usage for devices \texttt{AGX} (Jetson AGX Xavier), \texttt{MaxQ} (GTX 1080 MaxQ) and \texttt{RPI} (Raspberry Pi 400).

\textbf{Pytorch implementations.} Figures \ref{fig:esrmax}, \ref{fig:esrtm}, \ref{fig:esrtr} and \ref{fig:esrcnn} show the Python code to implement all proposed edge--SR models using the Pytorch tensor processing framework version 1.8. Model files and sample code are also available in \href{https://github.com/pnavarre/eSR}{https://github.com/pnavarre/eSR}.

{\small
\bibliographystyle{ieee_fullname}
\bibliography{bibliography}
}



\begin{figure*}
  \centering\vspace*{-.5cm}\hspace*{-1.7cm}
  \includegraphics[width=1.22\textwidth]{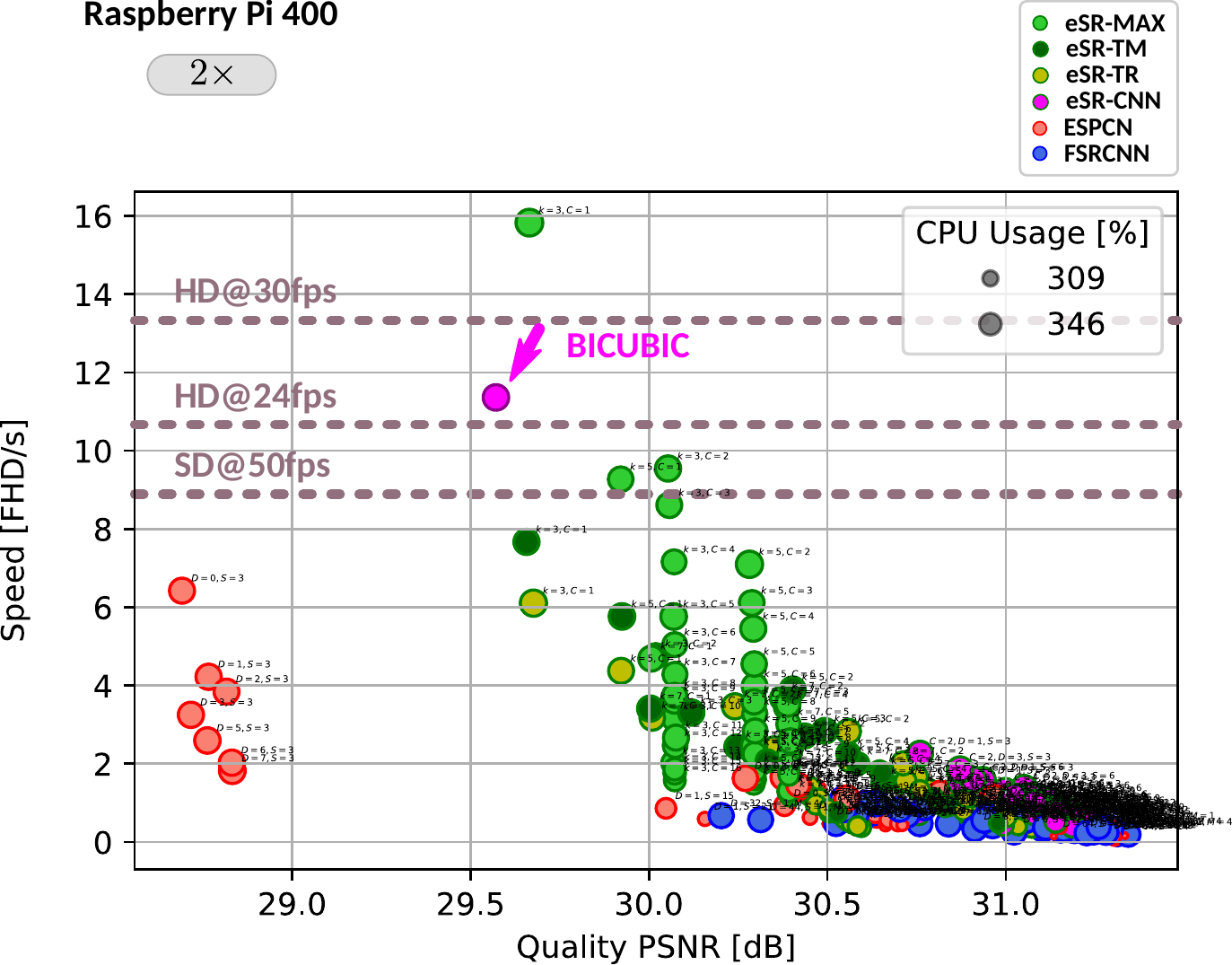}
  \caption{Speed of $2\times$ image SR models, measured in Full--HD pixels per second on a Raspberry Pi 400 CPU using 32--bit floating point precision, with respect to image quality, measure as PSNR in the BSDS--100 dataset. \label{fig:rpi_speed_vs_psnr_2x}}
\end{figure*}
\begin{figure*}
  \centering\vspace*{-.5cm}\hspace*{-1.7cm}
  \includegraphics[width=1.22\textwidth]{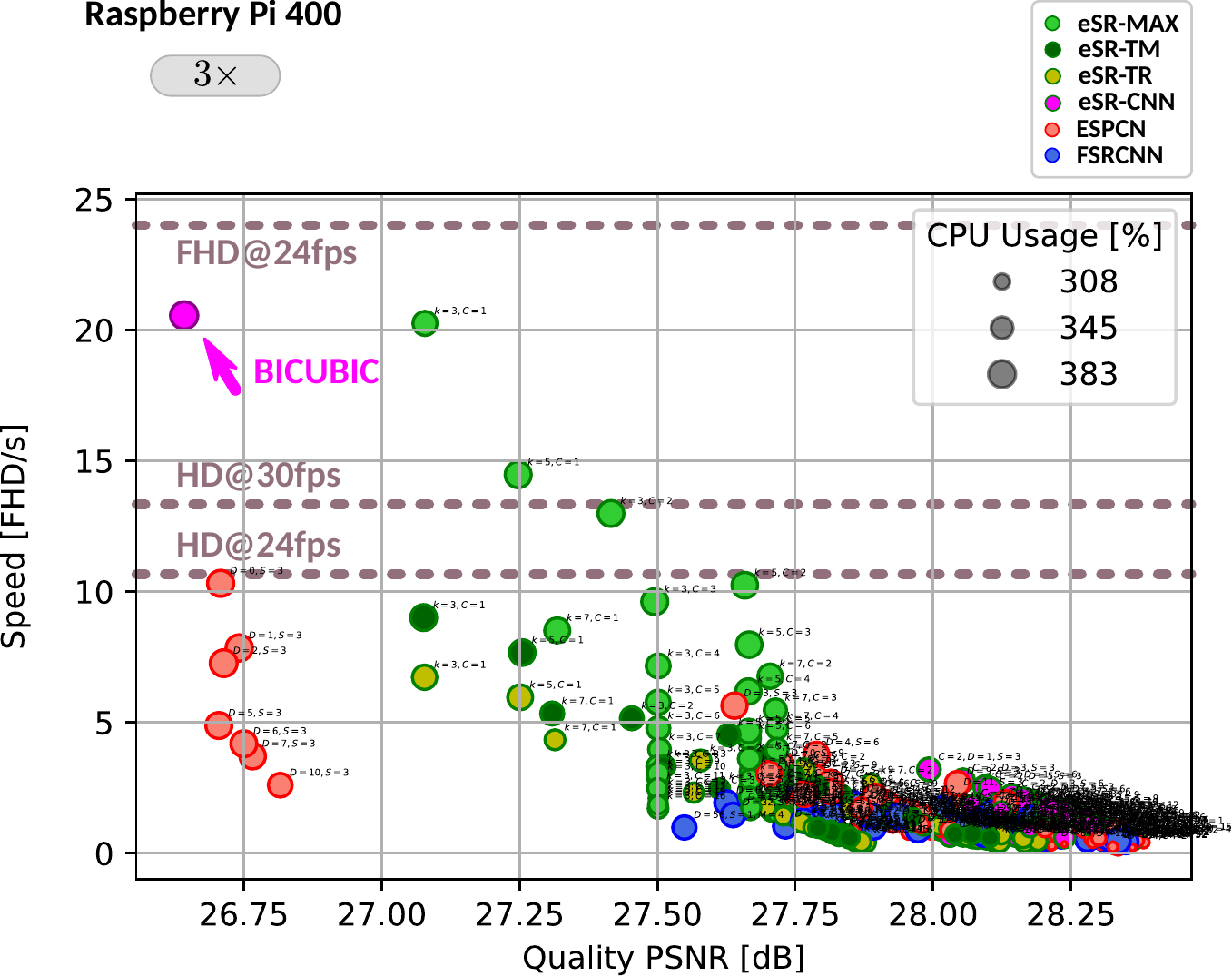}
  \caption{Speed of $3\times$ image SR models, measured in Full--HD pixels per second on a Raspberry Pi 400 CPU using 32--bit floating point precision, with respect to image quality, measure as PSNR in the BSDS--100 dataset. \label{fig:rpi_speed_vs_psnr_3x}}
\end{figure*}
\begin{figure*}
  \centering\vspace*{-.5cm}\hspace*{-1.7cm}
  \includegraphics[width=1.22\textwidth]{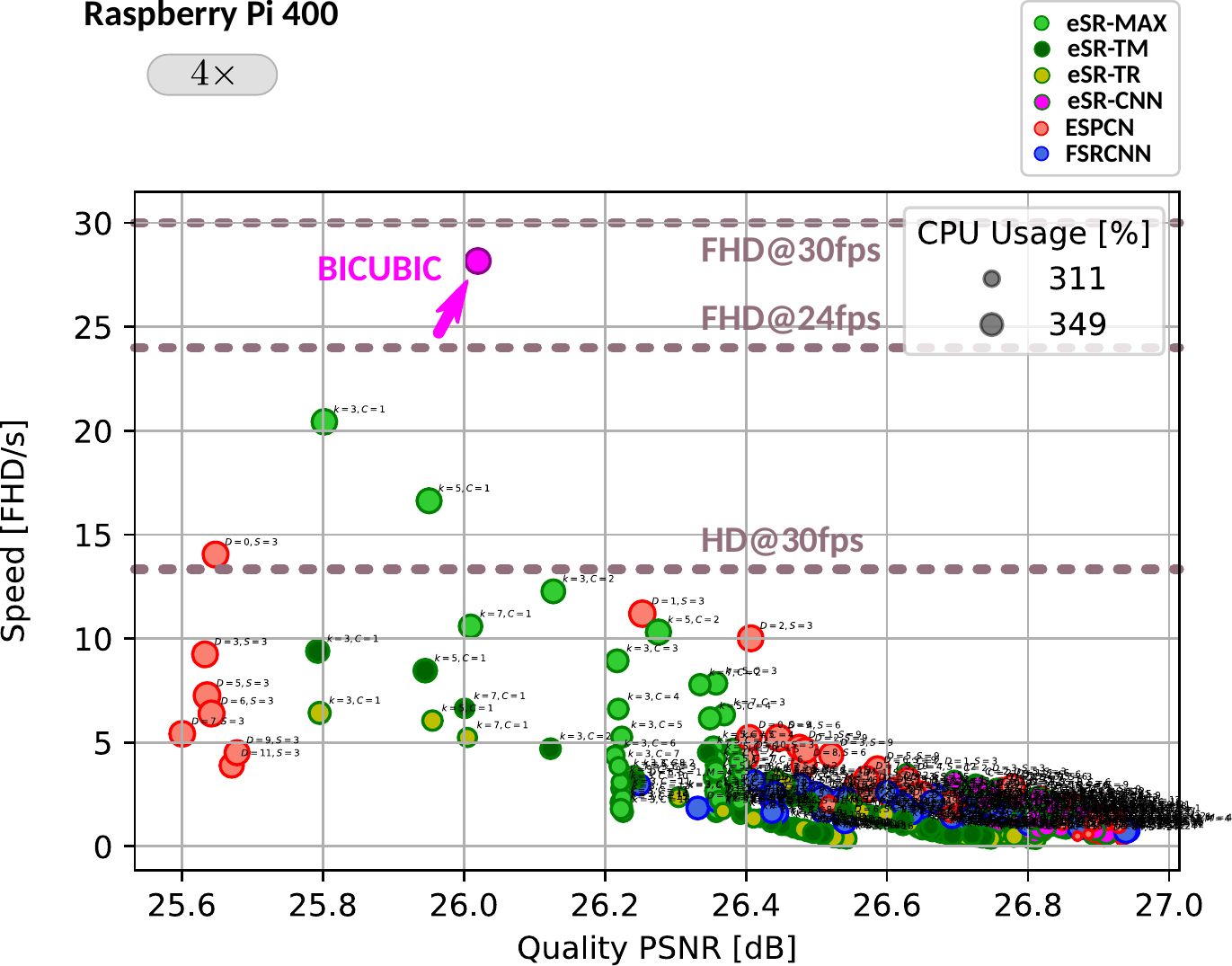}
  \caption{Speed of $4\times$ image SR models, measured in Full--HD pixels per second on a Raspberry Pi 400 CPU using 32--bit floating point precision, with respect to image quality, measure as PSNR in the BSDS--100 dataset. \label{fig:rpi_speed_vs_psnr_4x}}
\end{figure*}


\end{document}